\documentclass[journal]{IEEEtran}

\usepackage{amsmath}
\usepackage{amssymb}
\usepackage{amsfonts}
\usepackage{graphicx}
\usepackage{epsfig}
\usepackage{subfigure}
\usepackage{psfrag}
\usepackage{cite}
\usepackage{latexsym}
\usepackage{url}
\usepackage{color}
\usepackage{multirow}
\usepackage{mathtools}
\usepackage{bm}
\usepackage{booktabs}
\usepackage{algorithm}
\usepackage{algpseudocode}
\usepackage{bm}
\usepackage{verbatim}

\usepackage{verbatim}
\usepackage{indentfirst}
\usepackage{hyperref}
\hypersetup{
    colorlinks=true,
    linkcolor=blue,
    citecolor=blue,
    filecolor=blue,
    urlcolor=blue,}
\usepackage{multirow}
\usepackage{bbm}

\graphicspath{{fig/}}

\PassOptionsToPackage{bookmarks={false}}{hyperref}

\IEEEoverridecommandlockouts

\newtheorem{assumption}{Assumption}
\newtheorem{theorem}{\underline{Theorem}}
\newtheorem{lemma}{\underline{Lemma}}

\newtheorem{proposition}{\underline{Proposition}}

\newtheorem{remark}{\underline{Remark}}

\allowdisplaybreaks[4]

\begin{document}
\bstctlcite{IEEEexample:BSTcontrol}

\title{Closing the Generalization Gap in Parameter-efficient Federated Edge Learning}

\author{
Xinnong~Du, Zhonghao~Lyu,~\IEEEmembership{Member,~IEEE},
Xiaowen~Cao,~\IEEEmembership{Member,~IEEE},
Chunyang~Wen, Shuguang~Cui,~\IEEEmembership{Fellow,~IEEE}, 
and~Jie~Xu,~\IEEEmembership{Fellow,~IEEE}%
\thanks{
Xinnong Du, Shuguang Cui, and Jie Xu are with the School of Science and Engineering (SSE), the Shenzhen Future Network of Intelligence Institute (FNii-Shenzhen), and the Guangdong Provincial Key Laboratory of Future Networks of Intelligence, The Chinese University of Hong Kong (Shenzhen), Longgang, Shenzhen 518172, China (e-mail: xinnongdu@link.cuhk.edu.cn, shuguangcui@cuhk.edu.cn, xujie@cuhk.edu.cn).\\
\hspace*{1em}Zhonghao Lyu is with the School of Electrical Engineering and Computer Science, KTH Royal Institute of Technology, 114 28 Stockholm, Sweden (e-mail: lzhon@kth.se).\\
\hspace*{1em}Xiaowen Cao is with the College of Electronic and Information Engineering, Shenzhen University, and Guangdong Provincial Key Laboratory of Future Networks of Intelligence, The Chinese University of Hong Kong (Shenzhen), Longgang, Shenzhen 518172, China (email: caoxwen@szu.edu.cn). Xiaowen Cao is the corresponding author.\\
\hspace*{1em}Chunyang Wen is with University of Science and Technology of China (USTC), Hefei 230026, China (email: chywen@mail.ustc.edu.cn).}%
}

\date{2025/5/6}
\maketitle
\thispagestyle{empty}

\begin{abstract}
Federated edge learning (FEEL) provides a promising foundation for edge artificial intelligence (AI) by enabling collaborative model training while preserving data privacy. However, limited and heterogeneous local datasets, as well as resource-constrained deployment, severely degrade both model generalization and resource utilization, leading to a compromised learning performance. Therefore, we propose a parameter-efficient FEEL framework that jointly leverages model pruning and client selection to tackle such challenges. First, we derive an information-theoretic generalization statement that characterizes the discrepancy between training and testing function losses and embed it into the convergence analysis. It reveals that a larger local generalization statement can undermine the global convergence. Then, we formulate a generalization-aware average squared gradient norm bound minimization problem, by jointly optimizing the pruning ratios, client selection, and communication-computation resources under energy and delay constraints. Despite its non-convexity, the resulting mixed-integer problem is efficiently solved via an alternating optimization algorithm. Extensive experiments demonstrate that the proposed design achieves superior learning performance than state-of-the-art baselines, validating the effectiveness of coupling generalization-aware analysis with system-level optimization for efficient FEEL.

\end{abstract}

\section{Introduction}
\subsection{Background}
Driven by the deep integration of communication networks and artificial intelligence (AI), edge AI is expected to become a key application scenario for next-generation wireless systems \cite{Zhu2023}. However, in edge AI, the distributed data resources across edge environments are difficult to be fully exploited for effective model training \cite{Yuan2024}. To address this issue, federated edge learning (FEEL) has emerged as a promising paradigm that enables collaborative model training while preserving data privacy \cite{Zhu2023}.

Despite the benefits, the FEEL framework often faces the data isolation problem and suffers from overfitting, as locally trained models at distributed edge nodes tend to bias toward their own data, leading to degraded generalization performance \cite{Huang2024}. This issue becomes more severe in non-independent and identically distributed (non-IID) scenarios \cite{Lu2024}, where heterogeneous client data distributions result in slower convergence. This fundamentally limits the robustness and scalability of edge AI, underscoring the need to enhance model generalization in distributed learning.

\subsection{Related Works}
In FEEL, existing studies generally adopt three representative architectures, namely centralized, decentralized, and hierarchical architectures, corresponding to model aggregation at a central server~\cite{Zhang2024}, direct inter-client coordination~\cite{Kong2021}, and joint edge-cloud cooperation~\cite{Wang2023}, respectively. Within these architectures, achieving reliable learning performance in practical deployment over wireless edge networks remains challenging, due to the limited communication and computation capabilities at distributed edge nodes~\cite{Khowaja2021}, \cite{Cai2025}. To address this challenge, several studies have characterized learning performance in terms of the optimality gap through convergence analysis, and leveraged these insights to guide system-level optimization \cite{10422876,Pervej2025}. Specifically, some prior works have investigated the joint allocation of communication (e.g., bandwidth and transmit power) and computation resources (e.g., CPU frequency)~\cite{Zhu2021}, as well as training configurations (e.g., batch size and training rounds)~\cite{Park2021}, to minimize the optimality gap while reducing the learning delay and energy consumption. Moreover, to handle data heterogeneity, existing works have incorporated client selection and algorithm designs such as gradient correction to mitigate the impact of non-IID data distributions~\cite{Ruan2021}, \cite{Guo2025}, \cite{Zawad2025}. Meanwhile, beyond optimizing learning performance, other studies have further studied the minimization of overall energy consumption or learning latency to further enhance system efficiency, while ensuring the learning performance~\cite{Liu2024}, \cite{Yang2021}.

In addition to optimizing the convergence speed or energy efficiency, another line of work focuses on enhancing the generalization performance of FEEL models via techniques, such as data sharing \cite{Hu2023}, knowledge transferring \cite{Meng2025}, and mathematical analysis \cite{Li2024}. In particular, the authors in \cite{Hu2023} have used the data sharing and global datasets synthesizing to achieve more balanced data distributions across clients, thereby enabling unbiased local learning and mitigating the detrimental effects of statistical heterogeneity. \cite{Meng2025} has leveraged knowledge transfer, such as the aggregated global model outputs or the average local predictions, to refine local update directions and enhance cross-client output consistency, leading to improved model alignment. In addition, other works have adjusted aggregation strategies or adopted two-stage learning frameworks that sequentially perform global training and local fine-tuning, effectively strengthening the generalization performance \cite{Lyu2025}, \cite{Luo2021}. However, such data sharing and knowledge transfer strategies compromise the privacy-preserving capability of FEEL. Meanwhile, the additional transmission of the sharing data exacerbates the challenge of constrained edge resources. Moreover, recent works have employed information-theoretic frameworks to derive generalization analysis by quantifying the mutual information between data labels and extracted features~\cite{Li2024}, \cite{Barnes2022GenBounds} or by establishing tighter probably approximately correct (PAC) Bayesian generalization bounds under non-IID settings~\cite{Boroujeni2025}. Although these studies provide interpretable theoretical analyses of generalization, they do not consider how such insights can be incorporated into system design aspects.

On the other hand, with the increasing scale of model size, deploying AI models on resource-limited edge devices becomes challenging \cite{Ni2025}. Model compression techniques such as sparsification \cite{Deng2025}, quantization \cite{Lang2023}, 
and pruning \cite{lyu2025largermerrierefficientlarge,Yi2024,Liu20222,Molchanov2019,Pei2025} provide an effective solution for alleviating the resource bottlenecks. 
Among them, model pruning has been widely used for designing efficient lightweight FEEL recently. Existing works mainly focus on the characterization of parameter importance and the adjustment of pruning ratio. Specifically, some works have estimated parameter redundancy based on local accuracy gains to determine the pruning ratio \cite{Yi2024}, while others have jointly optimized pruning ratios with respect to (w.r.t.) the computational and communication resources of edge devices \cite{Liu20222}. Once the pruning ratio is determined, the clients prune less important parameters to achieve lightweight training, where the importance is represented by gradient variations \cite{Molchanov2019} or model weight magnitudes \cite{Pei2025}. However, existing lightweight FEEL frameworks mainly focus on reducing the communication and computation costs through isolated compression mechanisms, while overlooking their coupling with learning dynamics and generalization behavior.

\subsection{Motivations and Contributions}
Considering the above issues, it is essential to accurately analyze the generalization behavior of FEEL and incorporate such analysis into its parameter-efficient design. However, characterizing such behavior is fundamentally challenging. First, there is no clear analytical way to describe the mismatch between local training data and the target task, as each client observes only a limited and biased dataset. Consequently, it is non-trivial to establish a practical measure of how well a client’s local update can generalize to unseen data. Second, incorporating generalization into convergence analysis to reflect the model’s reliability on unseen tasks is also challenging, as it requires capturing the effects of key system parameters such as client selection and data imbalance. Finally, the deployment of generalization-aware and parameter-efficient FEEL at the network edge introduces another layer of complexity. Effectively exploiting heterogeneous and constrained resources at edge clients is highly challenging. Moreover, parameter-efficient strategies are closely coupled to the resource management schemes. This coupling makes it challenging to balance learning performance, energy efficiency, and delay. These challenges have not been well investigated in existing works, thus motivate our work. 

In this paper, we propose a parameter-efficient FEEL system to enhance generalization performance in heterogeneous and resource-constrained edge environments. We derive a novel learning performance bound by analyzing the generalization gap, which quantitatively captures the discrepancy between training and testing behaviors. Based on this analysis, we jointly optimize system resources, client selection, and model pruning ratios to accelerate the learning convergence. The main contributions are summarized as follows.
\begin{itemize}
    \item \textbf{Generalization analysis:} We derive a novel theoretical generalization statement, grounded in information-theoretic principles, that quantifies the deviation between local training and testing distributions. We incorporate the derived generalization statement into the convergence analysis and couple it with the client selection indicator to mitigate the impact of data heterogeneity on generalization. It is noteworthy that when the local training distribution of the client is more closely aligned with the sampled testing distribution, the gap between its training and test losses is reduced. This observation sheds light on how to schedule participating devices to improve generalization performance.
    \item \textbf{Joint optimization framework:} Building on the convergence and generalization analysis, we formulate an optimization problem that minimizes the generalization-aware average squared gradient norm bound, by jointly optimizing client participation, model pruning ratios, and communication and computation parameters under the constraints on overall system energy consumption and delay. Subsequently, we design an efficient algorithm to tackle this complex non-convex problem by leveraging alternating optimization based on successive convex approximation (SCA).
    \item \textbf{Simulation evaluation:} We conduct extensive simulations to evaluate the performance of the proposed design, where data heterogeneity across clients is simulated using a Dirichlet-based non-IID setting. The results demonstrate that, under given energy and latency constraints, our design significantly achieves higher test accuracy compared with other benchmark schemes. This improvement stems from the system-level joint resource optimization guided by generalization analysis.
\end{itemize}

The remainder of this paper is organized as follows. Section~\ref{sec:system_model} describes the proposed FEEL system model. Section~\ref{sec:analysis} provides convergence and generalization analysis. Section~\ref{sec:framework} presents the joint optimization framework that integrates generalization analysis with resource constraints. Section~\ref{sec:experiments} provides numerical results on benchmark datasets under various non-IID settings. Finally, Section~\ref{sec:conclusion} concludes the paper.

\section{System Model}\label{sec:system_model}
\begin{figure}[t]
	\centering
	\includegraphics[width=88mm]{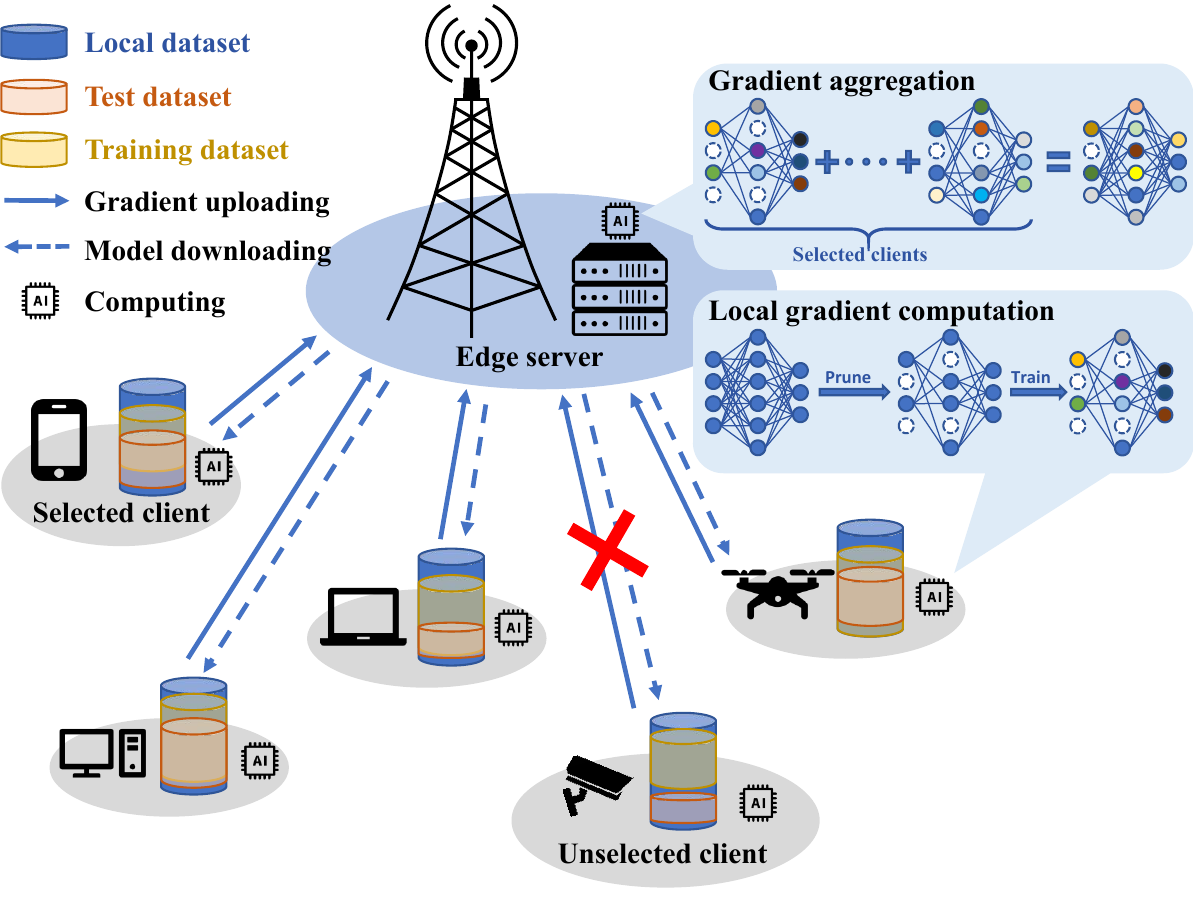}
	\caption{Illustration of the considered FEEL system over wireless communication networks with model pruning.}
	\label{fig1:env}
\end{figure}

\begin{figure}[t]
	\centering
	\includegraphics[width=89mm]{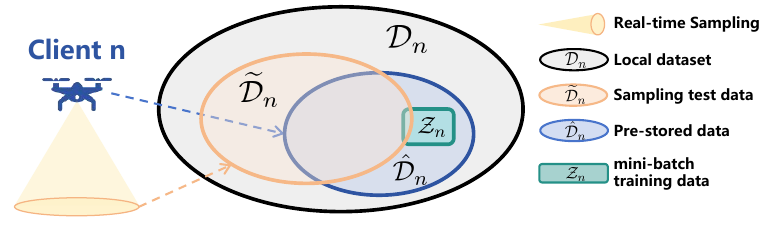}
	\caption{Hierarchical structure of the local dataset at client $n$.}
	\label{fig1:data}
\end{figure}

We consider a FEEL system consisting of an edge server and multiple clients with a set of $\mathcal{N}=\{1,\dots,N\}$ as shown in Fig.~\ref{fig1:env}. Suppose that client $n\in\mathcal{N}$ has a local dataset $\mathcal{D}_n$ comprising the task-oriented data, as illustrated in Fig.~\ref{fig1:data}. Denote $\mathcal{D}_n=\{ (x_{n,i},y_{n,i} )\}_{i=1}^{D_n}$ with size of $D_n$, where $x_{n,i}$ and $y_{n,i}$ represent the data sample and its corresponding ground-truth label, respectively. It is assumed that client $n$ has a pre-stored training dataset $\hat{\mathcal{D}}_n$ with size of $\hat{D}_n$ and a testing dataset $\tilde{\mathcal{D}}_n$ with size of $\tilde{D}_n$. To address the resource constraints in distributed edge AI scenarios, we adopt a parameter-efficient training strategy in the following.

\subsection{Parameter-efficient FEEL}
The FEEL enables the server and clients to collaboratively learn a shared model, represented by the parameter vector $\boldsymbol{\omega}\in\mathbb{R}^M$, where $M$ is the model size. In general, the FEEL algorithm aims at minimizing the global loss to obtain optimized model parameters, i.e.,
\begin{align}
    \boldsymbol{\omega}^{*}=\arg\min_{\boldsymbol{\omega}}\mathcal{L}(\boldsymbol{\omega},\hat{\mathcal{D}}),
\end{align}
where $\mathcal{L}(\boldsymbol{\omega},\hat{\mathcal{D}})$ denotes the loss function w.r.t. model parameters $\boldsymbol{\omega}$ over the entire training dataset $\hat{\mathcal{D}}\triangleq\cup^N_{n=1}\hat{\mathcal{D}}_n$. To solve this problem, we adopt the federated stochastic gradient descent (FedSGD) framework~\cite{McMahan2017} over training rounds with a set of $\mathcal{S} = \{0,1,\ldots,S\}$, during which client selection and model pruning are incorporated to improve the parameter efficiency. 

First, we introduce the client selection strategy. Denote $a_n^{(s)}$ as the binary selection indicator, where $a_n^{(s)} = 1$ if client $n$ is selected in round $s\in\mathcal{S}$. We also assume that the set of selected clients is denoted as $\tilde{\mathcal{N}}^{(s)}$ with size of $\tilde{N}^{(s)}$.

Then, we introduce the model pruning. We define the pruning ratio of client $n$ at round $s$ as $\lambda_n^{(s)}$, which is the ratio of the pruned model size $\tilde{M}_n^{(s)}$ to the full model size $M$, i.e., 
\begin{align}
	\lambda_n^{(s)}=\frac{\tilde{M}_n^{(s)}}{M},\forall n,s.\label{eq2}
\end{align}
We denote $\boldsymbol{Q}_n^{(s)}=\{Q_{n,m}^{(s)}\}_{m=1}^M$ as the importance score of model weights, which is measured by the squared error in the loss caused by the removal of the corresponding weight \cite{Molchanov2019}. For each parameter $m$, the corresponding importance score at client $n$ in round $s$ is given by
\begin{align}\label{imp}
	Q_{n,m}^{(s)}=&(\bar{\mathcal{L}}_n^{(s)}(\boldsymbol{\omega}_n^{(s)},\hat{\mathcal{D}}_n)-\bar{\mathcal{L}}_n^{(s)}(\boldsymbol{\omega}_n^{(s)}\vert_{\rho_{n,m}^{(s)}=0},\hat{\mathcal{D}}_n))^2,\forall m,n,s,
\end{align}
where $\boldsymbol{\omega}_n^{(s)}\vert_{\rho_{n,m}^{(s)}=0}$ is the unpruned model parameter vector $\boldsymbol{\omega}_n^{(s)}$ with the weight $\rho_{n,m}^{(s)}$ setting to be zero and $\bar{\mathcal{L}}_n^{(s)}(\boldsymbol{\omega}_n^{(s)},\hat{\mathcal{D}}_n)$ is the local loss function for client $n$ in round $s$. However, the evaluation in (\ref{imp}) is with high computation complexity. Denote $v_m^{(s-1)}$ as the global gradient of specific model weight $\rho_{n,m}^{(s-1)}$. To reduce the computational overhead, we adopt the first-order Taylor approximation as a surrogate~\cite{Liu20222}. Specifically, by expanding $\bar{\mathcal{L}}_n^{(s)}(\boldsymbol{\omega}_n^{(s)}\vert_{\rho_{n,m}^{(s)}=0},\hat{\mathcal{D}}_n)$, we have $\bar{\mathcal{L}}_n^{(s)}(\boldsymbol{\omega}_n^{(s)}\vert_{\rho_{n,m}^{(s)}=0},\hat{\mathcal{D}}_n)\approx \bar{\mathcal{L}}_n^{(s)}(\boldsymbol{\omega}_n^{(s)},\hat{\mathcal{D}}_n)- v_m^{(s-1)}\rho_{n,m}^{(s-1)}$. Substituting this approximation into (\ref{imp}) yields the first-order importance estimation
\begin{align}\label{importance_score}
	Q_{n,m}^{(s)}=(v_m^{(s-1)}\rho_{n,m}^{(s-1)})^2, \forall m=1,\cdots,M.
\end{align}

With client selection and model pruning, the FEEL is implemented as follows. At the beginning of each round, selected clients compute $\boldsymbol{Q}_n^{(s)}$ from the received global gradient and its local parameters through (\ref{importance_score}). Each selected client prunes its local model by removing weights of low importance according to $\boldsymbol{Q}_n^{(s)}$ and $\lambda_n^{(s)}$, and then obtain the pruned model $\tilde{\boldsymbol{\omega}}_n^{(s)}$. Denote $l(\tilde{\boldsymbol{\omega}}_n^{(s)};x_{n,i},y_{n,i})$ as the loss function of $\tilde{\boldsymbol{\omega}}_n^{(s)}$ on data point $(x_{n,i},y_{n,i})$. Then, the selected client $n$ randomly generates a mini-batch $\mathcal{Z}_n^{(s)}$ from $\hat{\mathcal{D}}_n$ with size of $Z_n^{(s)}\geq1$ in round $s$ to compute the local gradient, i.e.,
\begin{align}
    \nabla\bar{\mathcal{L}}_n^{(s)}(\tilde{\boldsymbol{\omega}}_n^{(s)},\mathcal{Z}_n^{(s)})=\frac{1}{Z_n^{(s)}}\sum_{i=1}^{Z_n^{(s)}}\nabla l(\tilde{\boldsymbol{\omega}}_n^{(s)};x_{n,i},y_{n,i}),\forall n,s. \!\!
\end{align}
Next, all selected clients upload their local gradients and the corresponding pruning indices to the server. To clearly define the global gradient, we use $\tilde{\boldsymbol{\omega}}^{(s)}$ to denote the global pruned model obtained by aggregating the selected local pruned models. Then, the server aggregates local gradients to obtain the global gradient over the entire training batch $\mathcal{Z}^{(s)}\triangleq\cup_{n\in\tilde{\mathcal{N}}^{(s)}}\mathcal{Z}_n^{(s)}$:
\begin{align}
	\nabla\mathcal{L}^{(s)}(\tilde{\boldsymbol{\omega}}^{(s)},\mathcal{Z}^{(s)})=\frac{1}{\tilde{N}^{(s)}}\sum_{n\in\tilde{\mathcal{N}}^{(s)}}\nabla \bar{\mathcal{L}}_n^{(s)}(\boldsymbol{\tilde{\omega}}_n^{(s)},\mathcal{Z}_n^{(s)}),\forall n,s,
\end{align}
based on which the server can update the global model as
\begin{align}
    \boldsymbol{\omega}^{(s+1)}=\boldsymbol{\omega}^{(s)}-\eta\nabla\mathcal{L}^{(s)}(\tilde{\boldsymbol{\omega}}^{(s)},\mathcal{Z}^{(s)}),
\end{align}
where $\eta$ denotes the learning rate. Finally, the server sends the global gradient back to all clients for local model update and importance score computation, while the pruning ratios are sent to the selected clients to start the next round. Notably, client selection and pruning ratio are determined by the server through the optimization process discussed in Section \ref{sec:framework}.

\subsection{Wireless Communication Model}
We then introduce the uplink communication model. All clients transmit their local gradients by using the technique of frequency-division multiple access (FDMA). Denote $p_n^{(s)}$ as the transmit power of client $n$ in round $s$. Then, the uplink rate (in bits-per-second (bps)) for client $n$ is
\begin{align}
	r^{(s)}_n\big(p_n^{(s)}\big)=c_n\mbox{log}_2\left(1+\frac{p^{(s)}_n{h^{(s)}_n}}{c_n U^{(s)}_0}\right), \forall n,s ,
\end{align}
where $c_n$ denotes the bandwidth allocated to client $n$, $h^{(s)}_n$ is the channel power gain from client $n$ to the server, and $U_0^{(s)}$ denotes the power spectral density (PSD) of the additive white Gaussian noise (AWGN) at the server in round $s$. 

Next, we introduce the downlink communication model. The server adopts a multicast transmission strategy, enabling the dissemination of identical data streams to multiple clients. Denote $\hat{p}$ as the transmit power of the server and $\hat{h}^{(s)}_n$ as the channel power gain from the server to client $n$ in round $s$. Then, the achievable downlink communication rate is
\begin{align}
	\hat{r}^{(s)}_n=\hat{c}\mbox{log}_2\left(1+\frac{\hat{p}{\hat{h}^{(s)}_n}}{\hat{c}U^{(s)}_{0,n}}\right), \forall n,s,
\end{align}
where $\hat{c}$ is the bandwidth for broadcasting the global gradient and $U_{0,n}^{(s)}$ is the PSD of AWGN at client $n$ in round $s$.

\subsection{System Delay and Energy Consumption}
\subsubsection{System Delay}
In general, the system delay in FEEL arises from both computation and communication. First, we analyze the computation delay, which primarily stems from local training, while we assume that the time for global aggregation is negligible. Reducing neurons or connections decreases the model size by eliminating operations associated with pruned nodes, such as multiplication, addition, and activation, thereby significantly lowering the needed number of floating-point operations (FLOPs). Moreover, the reduction in computational load is assumed to scale proportionally with the number of FLOPs~\cite{Liu20222}. Denote $f_n^{(s)}$ as the processor clock frequency of client $n$ in round $s$. Then, the computation delay at client $n$ is
\begin{align}
	\tau^{(s)}_n(\lambda_n^{(s)},f_n^{(s)})=\frac{(1-\lambda_n^{(s)})Z_n^{(s)}e_n}{f^{(s)}_nq_n},\forall n,s,
\end{align}
where $e_n$ denotes the number of FLOPs for computing the complete gradient of one data sample for client $n$, $(1-\lambda_n^{(s)})e_n$ corresponds to that for computing the pruned gradient, and $q_n$ denotes the number of FLOPs per clock cycle of the processor. 

Next, we analyze the communication delay. Denote $H_n^{(s)}$ as the data size (in bits) of the unpruned model gradients. Accordingly, we have $(1-\lambda_n^{(s)})H_n^{(s)}$ as the data size to be transmitted after pruning. Then, the communication delay of client $n$ in round $s$ is
\begin{align}
    \hat{\tau}^{(s)}_n(\lambda_n^{(s)},p_n^{(s)})=\frac{(1-\lambda_n^{(s)})H_n^{(s)}}{r_n^{(s)}(p_n^{(s)})}+\frac{H_n^{(s)}}{\hat{r}^{(s)}_n}.
\end{align}

Finally, the overall delay refers to the cumulative maximum latency among the selected clients over the $S$ rounds, i.e.,
\begin{align}
	&T\big(\{a_n^{(s)},\lambda_n^{(s)},p_n^{(s)},f_n^{(s)}\}\big) \nonumber\\
    &=\sum_{s=0}^{S}\max_{n\in \mathcal{N}}\big(a_n^{(s)}\big(\tau^{(s)}_n(\lambda_n^{(s)},f_n^{(s)})+\hat{\tau}^{(s)}_n(\lambda_n^{(s)},p_n^{(s)})\big)\big) \nonumber\\
    &=\sum_{s=0}^{S}\max_{n\in\mathcal{N}}\bigg(a_n^{(s)}\big(\frac{(1-\lambda_n^{(s)})Z_n^{(s)}e_n}{f^{(s)}_nq_n}+\frac{(1-\lambda_n^{(s)})H_n^{(s)}}{r_n^{(s)}(p_n^{(s)})} \nonumber\\
    &\quad\quad\quad\quad\quad+\frac{H_n^{(s)}}{\hat{r}^{(s)}_n})\bigg).
\end{align}

\subsubsection{Energy Consumption}
The energy consumption primarily arises from computation and communication. 
First, we analyze the energy consumption of local training, which is proportional to the number of FLOPs through model pruning. Accordingly, an approximately proportional relationship to the retained model size is applied~\cite{Guan2025}. Then, the computation energy consumption of client $n$ in round $s$ is
\begin{align}
    \tilde{E}_n^{(s)}(\lambda_n^{(s)},f_n^{(s)}) = (1-\lambda_n^{(s)})\kappa_n\varpi_n (f_n^{(s)})^2 \frac{Z_n^{(s)} e_n}{q_n},
\end{align}
where $\kappa_n$ denotes the power usage effectiveness (PUE) of client $n$, and $\varpi_n$ represents the effective switched capacitance coefficient determined by the processor characteristics.

Then, the energy consumption for gradients uploading at client $n$ in round $s$ is
\begin{align}
    \hat{E}_n^{(s)}(\lambda_n^{(s)}, &p_n^{(s)}) =\frac{(1-\lambda_n^{(s)})p_n^{(s)}H_n^{(s)}}{r_n^{(s)}(p_n^{(s)})}.
\end{align}

Thus, the overall energy consumption in round $s$ consists of local computation, local gradients uploading, and global information broadcasting for all clients, i.e.,
\begin{align}
    &E\big(\{a_n^{(s)},\lambda_n^{(s)},p_n^{(s)},f_n^{(s)}\}\big) \nonumber\\
    &=\sum_{s=0}^{S}\bigg(\sum_{n=1}^{N} a_n^{(s)}(\tilde{E}_n^{(s)}(\lambda_n^{(s)},f_n^{(s)})+\hat{E}_n^{(s)}(\lambda_n^{(s)},p_n^{(s)})) \nonumber\\
    &\quad\quad\quad+\hat{p}\max_{n\in\mathcal{N}}\left\{\frac{H_n^{(s)}}{\hat{r}_n^{(s)}}\right\}\bigg) \nonumber\\
    &= \sum_{s=0}^{S}\bigg(\sum_{n=1}^{N} a^{(s)}_n\big((1-\lambda_n^{(s)})\kappa_n\varpi_n (f_n^{(s)})^2 \frac{Z_n^{(s)} e_n}{q_n} &\nonumber\\
    &\quad\quad\quad+p_n^{(s)} \frac{(1-\lambda_n^{(s)})H_n^{(s)}}{r_n^{(s)}(p_n^{(s)})}\big) + \hat{p}\max_{n\in\mathcal{N}}\left\{\frac{H_n^{(s)}}{\hat{r}_n^{(s)}}\right\}\bigg).
\end{align}

\section{Convergence Analysis}\label{sec:analysis}
\subsection{Assumptions for Convergence Analysis}
For the convergence analysis, we first introduce standard assumptions on the loss function and gradient estimation, as widely adopted in prior studies (see, e.g., \cite{Liu20222}).

\begin{assumption}[Smoothness]\label{Assump1} 
The gradient $\nabla\mathcal{L}(\boldsymbol{\omega}, \mathcal{D})$ is $\text{Lipschitz}$ continuous w.r.t. the model parameters. Accordingly, for any $\boldsymbol{\omega}$ and $\boldsymbol{\omega}'$, it holds that
\begin{align}\label{L_smooth}
	\Vert\nabla\mathcal{L}(\boldsymbol{\omega}, \mathcal{D})-\nabla\mathcal{L}(\boldsymbol{\omega}', \mathcal{D})\Vert\leq L\Vert\boldsymbol{\omega}-\boldsymbol{\omega}'\Vert,
\end{align}
where $L$ denotes the $\text{Lipschitz}$ constant.
The Hessian matrix of (\ref{L_smooth}) is presented as $\nabla^2\mathcal{L}(\boldsymbol{\omega},\mathcal{D})\preceq L\boldsymbol{I}$, where $\boldsymbol{I}$ is an identity matrix. Also we have
\begin{align}
	\mathcal{L}(\boldsymbol{\omega},\mathcal{D})-\mathcal{L}(\boldsymbol{\omega}',\mathcal{D})\leq\nabla\mathcal{L}(\boldsymbol{\omega}',\mathcal{D})^{\rm{T}}(\boldsymbol{\omega}-\boldsymbol{\omega}')+\frac{L}{2}\Vert\boldsymbol{\omega}-\boldsymbol{\omega}'\Vert^2,
\end{align}
where $\rm{T}$ represents the transpose operation.
\end{assumption}

From Assumption \ref{Assump1}, the gradients of loss function variations w.r.t. model parameters are bounded, ensuring smooth changes rather than abrupt fluctuations.

\begin{assumption}[Unbiased gradient]\label{Assump2} 
The global mini-batch stochastic gradient $G(\boldsymbol{\omega})$ is assumed to be an unbiased estimate of the full-batch gradient $\nabla\mathcal{L}(\boldsymbol{\omega},\hat{\mathcal{D}})$, expressed as
\begin{align}
	\mathbb{E}\left\{G(\boldsymbol{\omega})\right\}=\nabla\mathcal{L}(\boldsymbol{\omega},\hat{\mathcal{D}}),
\end{align}
where $\mathbb{E}\{\cdot\}$ denotes the statistical expectation.
\end{assumption}

\begin{assumption}[Bounded gradient and model]\label{Assump3} 
The second moments of the local mini-batch stochastic gradients $g(\boldsymbol{\omega})$ and model parameters are upper bounded by non-negative constants $A^2$ and $B^2$, respectively, i.e.,
\begin{align}
    &\mathbb{E}\left\{\Vert g(\boldsymbol{\omega})\Vert^2\right\}\leq A^2, \\
    &\mathbb{E}\left\{\Vert\boldsymbol{\omega}\Vert^2\right\}\leq B^2.
\end{align}
\end{assumption}
	
\begin{assumption}[Bounded pruning level]\label{Assump4}
 The expected squared difference between the model parameters before and after pruning is upper bounded by the pruning ratio $\lambda_n^{(s)}$ with the expected squared parameter norm~\cite{Molchanov2019}, i.e.,
\begin{align}
    \mathbb{E}\left\{ \Vert \boldsymbol{\omega}_n^{(s)}-\tilde{\boldsymbol{\omega}}_n^{(s)} \Vert^2 \right\}\leq\lambda_n^{(s)}\mathbb{E}\left\{ \Vert \boldsymbol{\omega}_n^{(s)} \Vert^2 \right\}.
\end{align}
\end{assumption}

\subsection{Convergence Analysis Based on Generalization Gap}
In the following, we analyze the convergence property of the proposed FEEL framework w.r.t. the generalization gap. Denote $\tilde{\mathcal{L}}(\boldsymbol{\omega}^{(s)},\tilde{\mathcal{D}})$ as the global loss function under global test dataset $\tilde{\mathcal{D}}$ in round $s$. First, we introduce the generalization performance. Specifically, the generalization gap is defined as a theoretical measure of the discrepancy between the training loss (empirical risk) and the test loss (population risk) \cite{Barnes2022GenBounds}, given by
\begin{align}
	\varphi^{(s)}\triangleq\mathcal{L}^{(s)}(\boldsymbol{\omega}^{(s)},\hat{\mathcal{D}})-\tilde{\mathcal{L}}^{(s)}(\boldsymbol{\omega}^{(s)},\tilde{\mathcal{D}}).
\end{align}

Next, we use the generalization gap and gradient decomposition during the FEEL training to characterize the upper bounds of model gradients on both the training and test sets. Specifically, given a data point $z$, we denote $\mathrm{p}(z\vert\hat{\mathcal{D}})$ and $\mathrm{p}(z\vert\tilde{\mathcal{D}})$ as the probability distributions over the training and test dataset, respectively, while $\mathrm{p}'(z\vert\hat{\mathcal{D}})$ denotes the probability of the least frequent data point in the training set. Denote $\mathrm{H}(\mathrm{p}(z\vert\tilde{\mathcal{D}}))$ as the entropy of test distribution, and $\mathrm{I}\big(\mathrm{p}(z\vert\hat{\mathcal{D}}),\mathrm{p}(z\vert\tilde{\mathcal{D}})\big)$ as the mutual information between the training and test distributions. With these preliminaries, we establish the following lemma to characterize the gradient discrepancy between the training and test sets.
\begin{lemma}
\label{lem1}
    For any model $\boldsymbol{\omega}$, the norm of the difference between the gradients evaluated over the training set $\hat{\mathcal{D}}$ and test set $\tilde{\mathcal{D}}$ is upper bounded by
	\begin{align}
		\Vert\nabla\mathcal{L}(\boldsymbol{\omega},\hat{\mathcal{D}})-\nabla\tilde{\mathcal{L}}(\boldsymbol{\omega},\tilde{\mathcal{D}})\Vert\leq\phi\Vert\nabla\mathcal{L}(\boldsymbol{\omega},\hat{\mathcal{D}})\Vert,
	\end{align}
    where $\phi$ is defined as the generalization statement, derived as $\phi=\bigg[\frac{(\hat{D}+\tilde{D})}{\mathrm{p}^{'}(z\vert\hat{\mathcal{D}})}\cdot \nonumber\bigg\vert\frac{\sqrt{2(\mathrm{H}(\mathrm{p}(z\vert\tilde{\mathcal{D}}))-\mathrm{I}\big(\mathrm{p}(z\vert\hat{\mathcal{D}}),\mathrm{p}(z\vert\tilde{\mathcal{D}})\big))}}{1-\tilde{D}\sqrt{2(\mathrm{H}(\mathrm{p}(z\vert\tilde{\mathcal{D}}))-\mathrm{I}\big(\mathrm{p}(z\vert\hat{\mathcal{D}}),\mathrm{p}(z\vert\tilde{\mathcal{D}})\big))}}\bigg\vert\bigg]$.
    
    \textit{Proof:} See Appendix \ref{appendix:lemma1}.
\end{lemma}

Next, building on the previous analysis, we introduce the following proposition that captures the generalization gap across each FEEL round.
\begin{proposition}
\label{lem2}
    The generalization gap between training rounds $s$ and $(s+1)$ is theoretically derived as
	\begin{align}
		\varphi^{(s+1)}&-\varphi^{(s)} \leq\frac{1}{2}\left(\eta^2+\left\vert\sum_{n=1}^{N}a_n^{(s)}\phi_n\right\vert^2\right)\mathbb{E}\left\{\Vert G(\tilde{\boldsymbol{\omega}}^{(s)})\Vert^2\right\},
	\end{align}
    where $\phi_n$ is the generalization statement at client $n$, derived as $\phi_n =\bigg[\frac{(\hat{D}_n+\tilde{D}_n)}{\mathrm{p}^{'}(z\vert\hat{\mathcal{D}}_n)}\cdot \nonumber\bigg\vert\frac{\sqrt{2(\mathrm{H}(\mathrm{p}(z\vert\tilde{\mathcal{D}}_n))-\mathrm{I}\big(\mathrm{p}(z\vert\hat{\mathcal{D}}_n),\mathrm{p}(z\vert\tilde{\mathcal{D}}_n)\big))}}{1-\tilde{D}_n\sqrt{2(\mathrm{H}(\mathrm{p}(z\vert\tilde{\mathcal{D}}_n))-\mathrm{I}\big(\mathrm{p}(z\vert\hat{\mathcal{D}}_n),\mathrm{p}(z\vert\tilde{\mathcal{D}}_n)\big))}}\bigg\vert\bigg]$.
    
    \textit{Proof:} See Appendix \ref{appendix:proposition1}.
\end{proposition}

\begin{remark}
Proposition~\ref{lem2} characterizes an upper bound on the variation of the generalization gap across consecutive training rounds, where the bound explicitly depends on the generalization statement $\phi_n$. We prefer reducing the generalization gap $\varphi^{(s+1)}$ in round $(s+1)$ compared with round $s$ by considering $\phi_n$, thereby achieving improved generalization. In particular, a smaller $\phi_n$ indicates that the local training distribution of a client is more consistent with the sampled testing distribution, thereby reducing the discrepancy between training and test losses. Consequently, prioritizing clients with smaller values of $\phi_n$ in the selection process can enhance the alignment between local updates and the global objective, thus improving model generalization performance. This observation highlights the importance of client selection for improving the generalization in FEEL.
\end{remark}  

Finally, building on Proposition~\ref{lem2}, we analyze the gradient convergence with generalization gap dynamics and client heterogeneity, as presented in Theorem~\ref{theo1}.
\begin{theorem}\label{theo1}
    With the FEEL model initialized at $\boldsymbol{\omega}^{(0)}$, satisfying Assumptions~\ref{Assump1}$\sim$\ref{Assump4}, with fixed learning rate $\eta$ and batch size $Z=Z_n^{(s)},\forall n,s$, the average squared gradient norm after $S$ training rounds has the following upper bound:
	\begin{align}\label{normb}
		\frac{1}{S+1}\sum\nolimits_{s=0}^S&\mathbb{E}\left\{\Vert \nabla\tilde{\mathcal{L}}(\tilde{\boldsymbol{\omega}}^{(s)},\tilde{\mathcal{D}})\Vert^2\right\}\leq\theta(\{a_n^{(s)},\lambda_n^{(s)}\}) \nonumber\\
		&\triangleq\alpha+\beta\sum_{s=0}^S\frac{1}{\sum_{n=1}^{N}a_n^{(s)}}\nonumber\\
        &+\sum_{s=0}^S\frac{\gamma_1\vert\sum_{n=1}^{N}a_n^{(s)}\phi_n\vert^2+\gamma_2\sum_{n=1}^{N}a_n^{(s)}\lambda_n^{(s)}}{\sum_{n=1}^{N}a_n^{(s)}},
	\end{align}
    where $\alpha=\frac{2(\mathcal{L}(\boldsymbol{\omega}^{(0)})-\mathcal{L}(\boldsymbol{\omega}^{*}))}{\eta(S+1)}$, $\beta=\frac{\eta^3A^2(L+1)}{Z(S+1)}$, $\gamma_1=\frac{\eta A^2}{Z(S+1)}$, and $\gamma_2=\frac{L^2B^2}{(S+1)}$. \\
    \textit{Proof:} See Appendix \ref{appendix:theorem1}.
\end{theorem}

\begin{remark}
The upper bound $\theta(\{a_n^{(s)},\lambda_n^{(s)}\})$ in Theorem~\ref{theo1} indicates that both the client selection and the pruning ratio directly affect the convergence behavior. Increasing the number of participated clients enhances update reliability, while smaller pruning ratios contribute to faster convergence. These jointly tighten the gradient bound and accelerate convergence, but heavier communication and computation from more clients and less-pruned models increase latency and energy consumption. Moreover, the system prioritizes the client with smaller $\phi_n$ that further accelerates convergence. Consequently, jointly optimizing $\{a_n^{(s)}\}$ and $\{\lambda_n^{(s)}\}$ alongside $\{p_n^{(s)}\}$ and $\{f_n^{(s)}\}$ is essential to balance learning (generalization) performance, energy consumption, and delay, thereby providing theoretical guidance for resource optimization.
\end{remark}

\section{Joint Resource Optimization} \label{sec:framework}
This section first formulates a joint optimization problem to enhance the convergence performance subject to system energy consumption and latency requirements. Then, we develop an efficient algorithm to solve the formulated problem.

\subsection{Problem Formulation}
We design the optimization problem to minimize the expected squared gradient norm bound $\theta(\{a_n^{(s)},\lambda_n^{(s)}\})$ in (\ref{normb}) by jointly optimizing the pruning ratios $\{\lambda_n^{(s)}\}$, client selection indicators $\{a_n^{(s)}\}$, computation frequencies $\{f_n^{(s)}\}$, and transmit powers $\{p_n^{(s)}\}$. Accordingly, the learning performance optimization problem is formulated as
\begin{subequations}\label{p1}
	\begin{align}
    \text{(P1)}:&\min_{\{\lambda_n^{(s)},a_n^{(s)},p_n^{(s)},f_n^{(s)}\}} \quad \theta(\{a_n^{(s)},\lambda_n^{(s)}\}) & \nonumber\\
		&\mbox{s.t.}\quad
		E(\{a_n^{(s)},\lambda_n^{(s)},p_n^{(s)},f_n^{(s)}\})\leq E_0,  & \label{P1-a}\\
		&\quad\quad~ T(\{a_n^{(s)},\lambda_n^{(s)},p_n^{(s)},f_n^{(s)}\})\leq T_0, &  \label{P1-b}\\
        &\quad\quad~ 0\leq\lambda_n^{(s)}\leq\lambda^{\mathrm{max}}, \forall n \in \mathcal{N}, & \label{P1-c}\\
		&\quad\quad~ 0\leq f_n^{(s)}\leq f^{\mathrm{max}}_n, \forall n \in \mathcal{N}, & \label{P1-d}\\
        &\quad\quad~ 0 \leq p_n^{(s)}\leq p^{\mathrm{max}}_n, \forall n \in \mathcal{N}, \label{P1-e}\\
		&\quad\quad~ a_n^{(s)} \in \left\{0,1\right\}, \forall n \in \mathcal{N}, & \label{P1-f}
	\end{align}
\end{subequations}
where $\lambda^{\mathrm{max}}$, $f^{\mathrm{max}}_n$, and $p^{\mathrm{max}}_n$ in \eqref{P1-c}$\sim$\eqref{P1-e} denote the maximum pruning ratio, clock frequency, and transmit power during parameter-efficient FEEL, respectively, $E_0$ and $T_0$ denote the overall energy consumption and system delay requirements. Appropriate adjustment of $E_0$ and $T_0$ ensures a trade-off between convergence performance, energy consumption, and system latency over the learning process. However, in problem (P1), the objective and constraint functions in (\ref{P1-a}) and (\ref{P1-b}) are non-convex, due to the tight coupling between $\{a_n^{(s)}\}$ and $\{\lambda_n^{(s)}\}$, as well as $\{p_n^{(s)}\}$ appearing in both the numerator and the denominator of (\ref{P1-a}). Moreover, due to the binary variable $\{a_n^{(s)}\}$, problem (P1) is a mixed-integer nonlinear program (MINLP), which is highly non-convex and hard to be optimally solved.

\renewcommand{\arraystretch}{1.0}  
\begin{table*}[t]
\centering
\caption{Experiment Setups}
\label{tab:exp_setup}
\begin{tabular}{p{5.5cm}|p{5cm}|p{5cm}}
\toprule
\textbf{Parameter} & \textbf{Model on MNIST} & \textbf{Model on CIFAR-10} \\
\midrule
Model/gradient size $H$ & 1.42 Mbit & 21.07 Mbit \\
Computation workload $e$ & 1.8 MFLOPs & 0.59 GFLOPs \\
Bandwidth of devices $c_n^U$ & 100 kHz &  2 MHz \\
PSD of AWGN $U_0^{(s)}$ & $3.98 \times 10^{-21}$ W/Hz & $3.98 \times 10^{-21}$ W/Hz \\
Maximum pruning ratio $\lambda^{\max}$ & 0.5 & 0.7 \\
Maximum clock frequency of clients $f_n^{\max}$ & 500 MHz & 2000 MHz \\
Maximum transmit power of clients $p_n^{\max}$ & 500 mW & 500 mW \\
FLOPs performed per clock cycle $q_n$ & 4 & 8 \\
The PUE of clients $\kappa_n$ & 1 & 1 \\
Power coefficient of clients $\{\varpi_n\}$ & 
\begin{minipage}[t]{\linewidth}\raggedright
$\{0.88, 0.84, 1.41, 1.33, 0.94, 1.37, 1.8,$\\
$1.01, 0.26, 0.96\} \times 10^{-27}$
\end{minipage} & 
\begin{minipage}[t]{\linewidth}\raggedright
$\{0.88, 0.84, 1.41, 1.33, 0.94, 1.37, 1.8,$\\
$1.01, 0.26, 0.96\} \times 10^{-28}$
\end{minipage} \\
\bottomrule
\end{tabular}
\end{table*}

\begin{figure*}[t]
    \centering
    \subfigure[$\sigma=1$]{
        \includegraphics[width=0.23\linewidth]{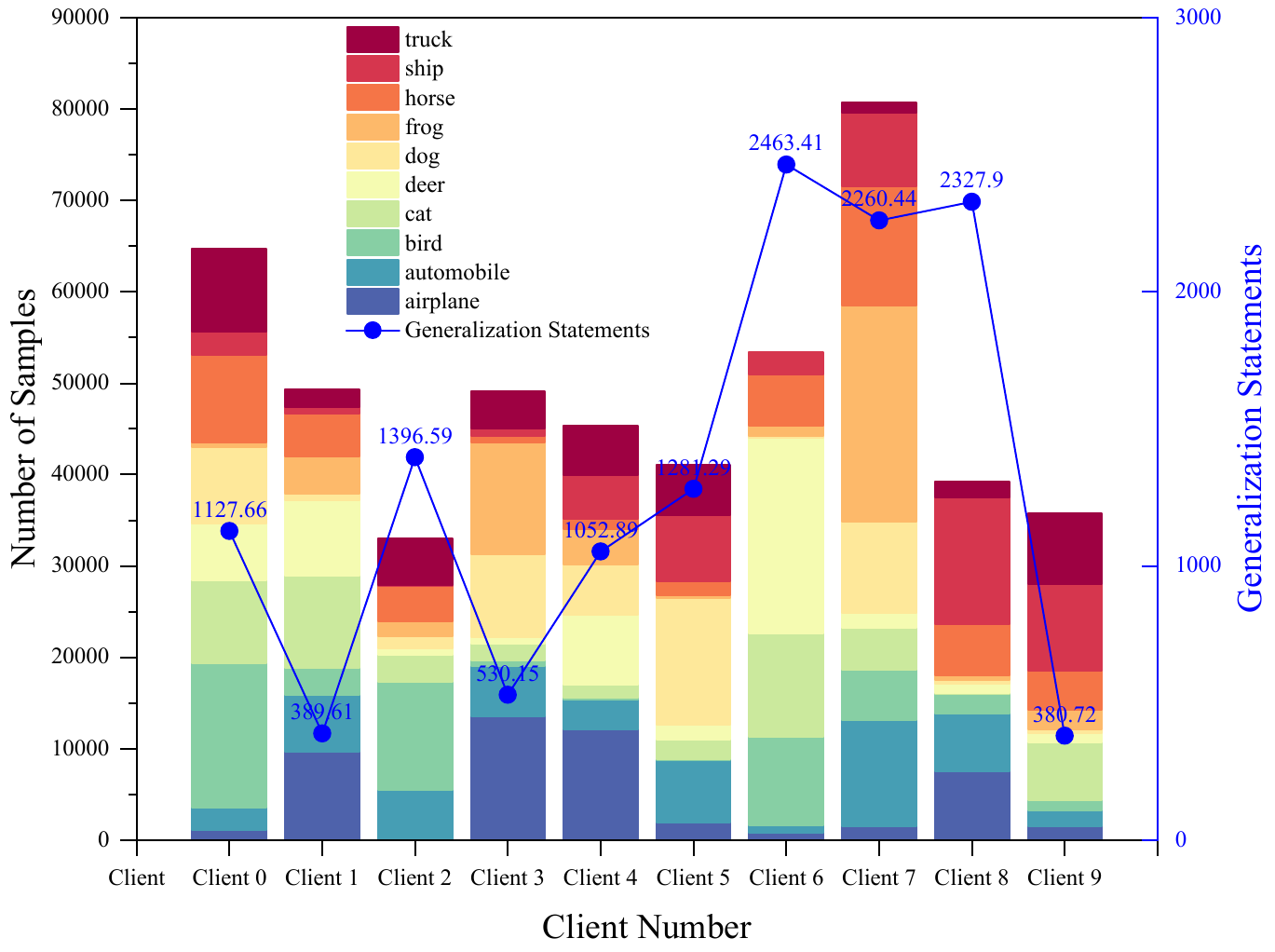}
        \label{5}}
    \subfigure[$\sigma=5$]{
        \includegraphics[width=0.23\linewidth]{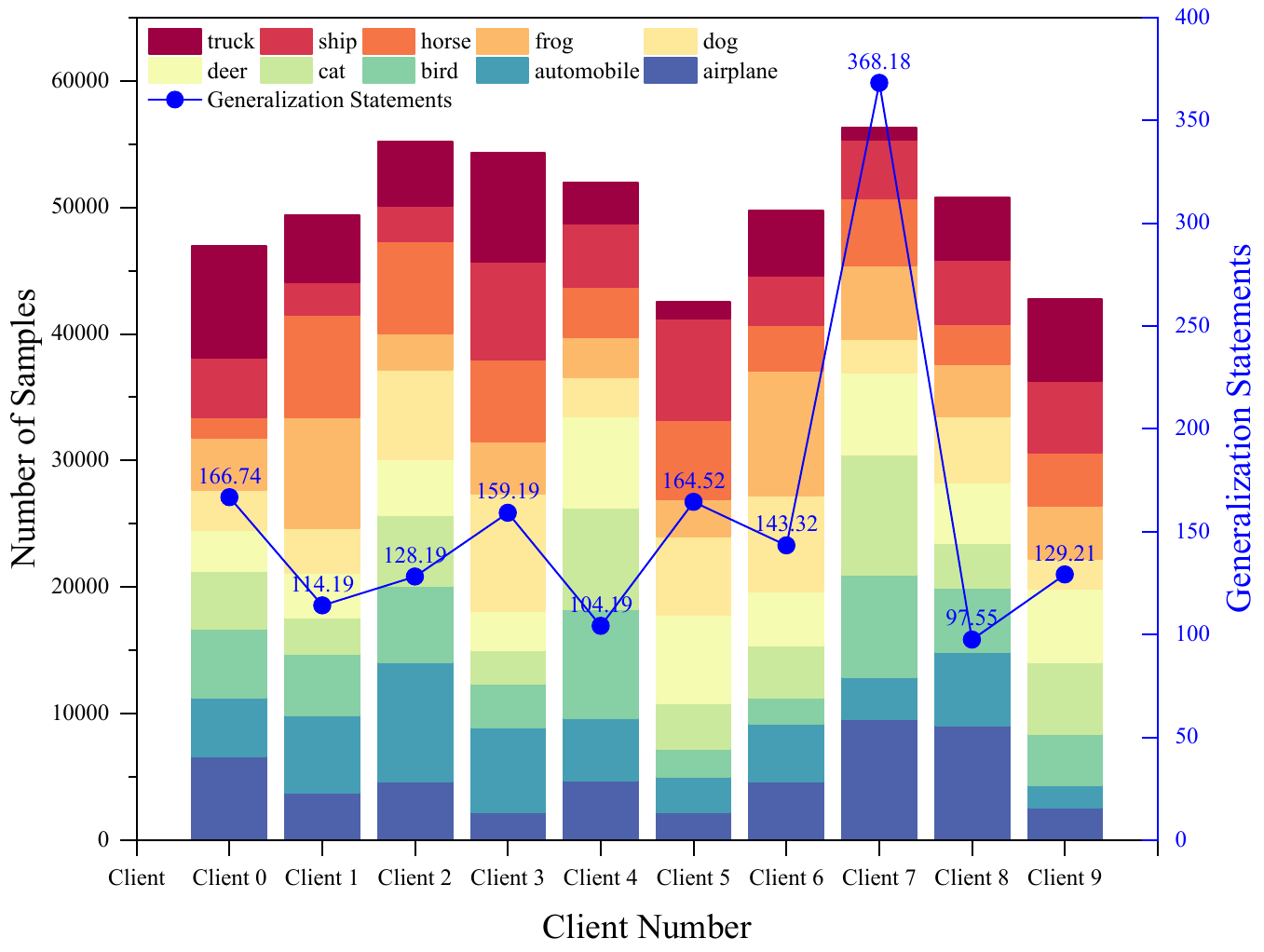}
        \label{6}}
    \subfigure[$\sigma=10$]{
        \includegraphics[width=0.23\linewidth]{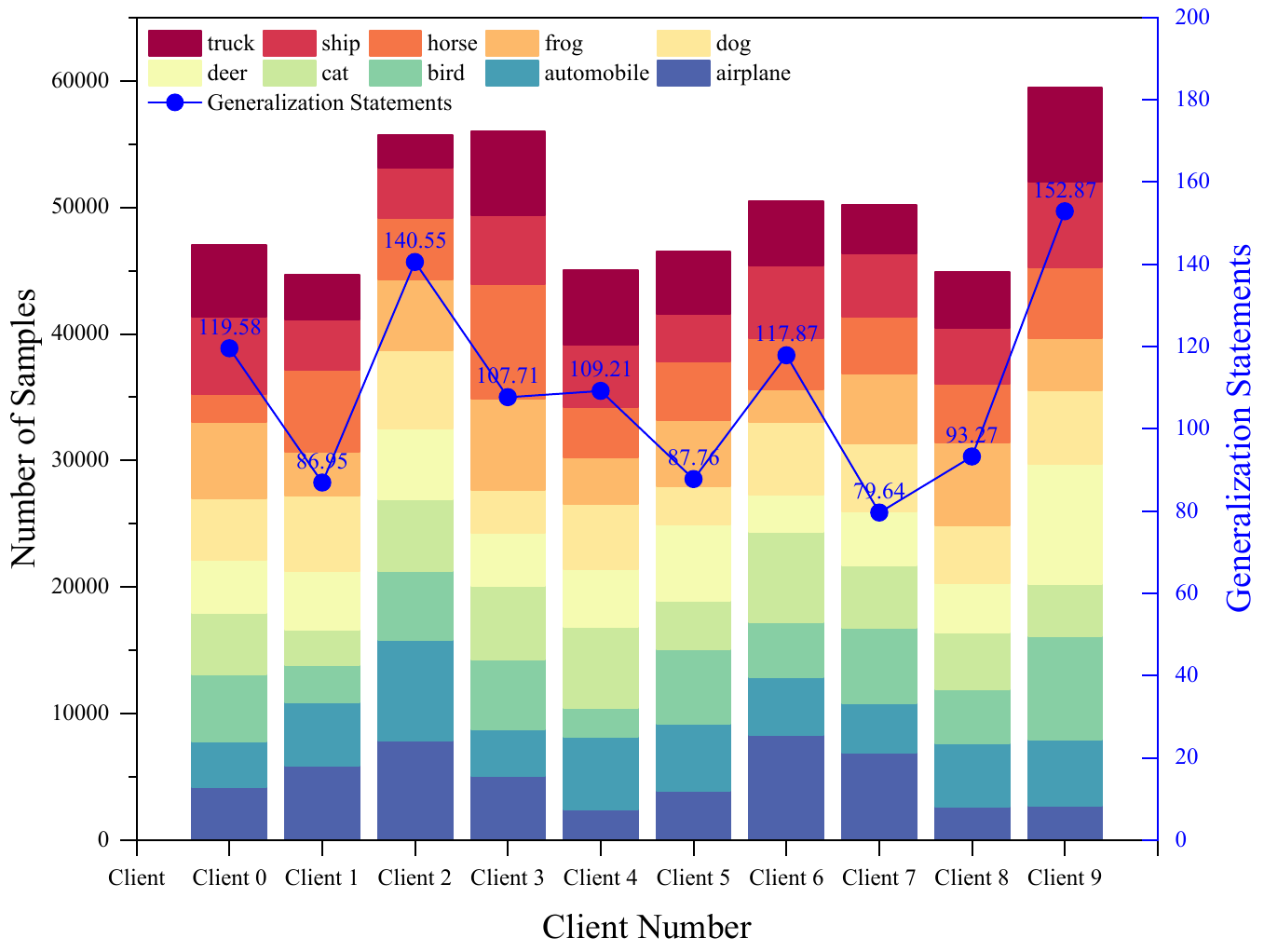}
        \label{7}}
    \subfigure[$\sigma=15$]{
        \includegraphics[width=0.23\linewidth]{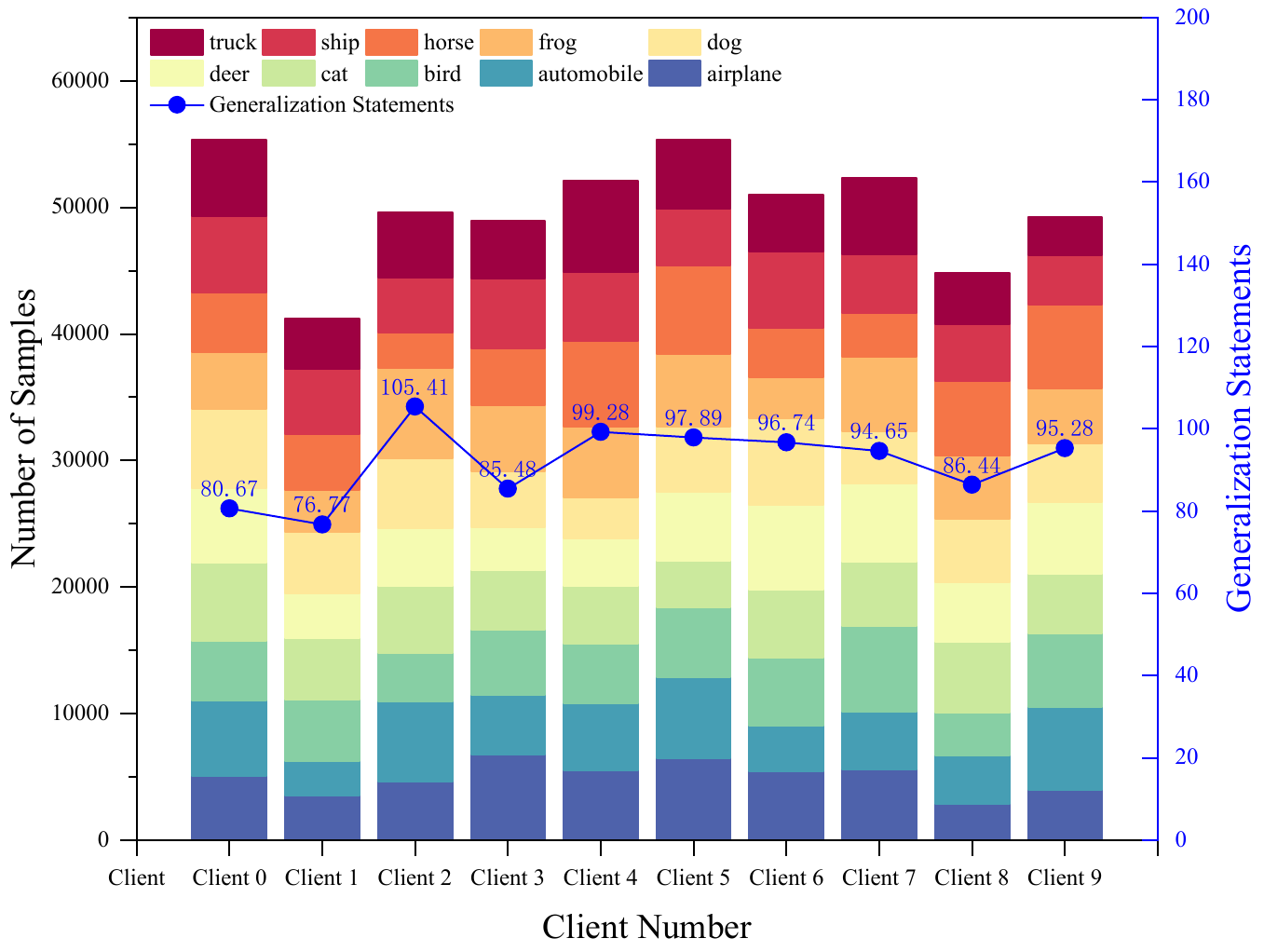}
        \label{8}}
    \caption{Impact of data heterogeneity on sample distributions and generalization statements with different values of $\sigma$.}
    \label{fig:data_dis}
\end{figure*}

\subsection{Proposed Solution to Problem (P1)}
We propose an efficient alternating optimization (AO) framework to tackle problem (P1). The framework decouples pruning ratios, selection indicators, and resource variables across the objective and constraints in an iterative manner.

\subsubsection{Optimization of System Resources}
First, with fixed $\{a_n^{(s)}\}$ and $\{\lambda_n^{(s)}\}$, we jointly optimize the communication resource allocation and computation frequency in the following problem (P2).
\begin{subequations}\label{P5}
	\begin{align}
		\text{(P2)}:& \min_{\{p_n^{(s)},f_n^{(s)}\}} \quad \theta(\{a_n^{(s)},\lambda_n^{(s)}\}) \nonumber\\
        &\mbox{s.t.}\quad
        \sum_{s=0}^{S}\bigg(\sum_{n=1}^{N} a^{(s)}_n\big((1-\lambda_n^{(s)})\kappa_n\varpi_n (f_n^{(s)})^2 \frac{Z_n^{(s)} e_n}{q_n} &\nonumber\\
        &\quad\quad~ +\frac{(1-\lambda_n^{(s)})p_n^{(s)}H_n^{(s)}}{c_n\mbox{log}_2(1+\frac{p^{(s)}_n{h^{(s)}_n}}{c_n U^{(s)}_0})}\big) + \hat{p}\max_{n\in\mathcal{N}}\big(\frac{H_n^{(s)}}{\hat{r}_n^{(s)}}\big)\bigg)\leq E_0, & \label{P2-a} \\
        &\quad\quad~ \sum_{s=0}^{S}\max_{n\in\mathcal{N}}\bigg(a_n^{(s)}\bigg(\frac{(1-\lambda_n^{(s)})Z_n^{(s)}e_n}{f^{(s)}_nq_n} &\nonumber\\
        &\quad\quad~ +\frac{(1-\lambda_n^{(s)})H_n^{(s)}}{c_n\mbox{log}_2(1+\frac{p^{(s)}_n{h^{(s)}_n}}{c_n U^{(s)}_0})}+\frac{H_n^{(s)}}{\hat{r}^{(s)}_n}\bigg)\bigg)\leq T_0, &  \label{P2-b}\\
		&\quad\quad~~(\ref{P1-d}), (\ref{P1-e}). & \nonumber
	\end{align}
\end{subequations}
Problem (P2) is still non-convex and hard to be optimally solved directly. We apply SCA to approximate the non-convex constraint into a convex counterpart. Specifically, we take the first-order Taylor expansion on the middle term of non-convex constraint \eqref{P2-a} at point $p_n^{(s)^{(k)}}$ in iteration $k\geq1$, i.e.,
	\begin{align}\label{SCA}
		\frac{p_n^{(s)}H_n^{(s)}}{c_n\mbox{log}_2(1+\frac{p^{(s)}_nh^{(s)}_n}{c_nU^{(s)}_0})}&\leq\frac{p^{(s)^{(k)}}_n H_n^{(s)}}{c_n\mbox{log}_2(1+\frac{p^{(s)^{(k)}}_n{h^{(s)}_n}}{c_n U^{(s)}_0})} \nonumber\\
		&+b_n^{(s)^{(k)}}(p_n^{(s)}-p_n^{(s)^{(k)}})\triangleq\xi^{(k)}(p_n^{(s)}),
	\end{align}
	where $b_n^{(s)^{(k)}}$ is the partial derivative over $p_n^{(s)^{(k)}}$, i.e.,
	\begin{align}
		&\bigg(\frac{\partial}{\partial p_n^{(s)^{(k)}}}\frac{p_n^{(s)^{(k)}}H_n^{(s)}}{c_n\mbox{log}_2(1+\frac{p^{(s)^{(k)}}_n h^{(s)}_n}{c_nU^{(s)}_0})}\bigg)=\frac{H_n^{(s)}}{c_n\mbox{log}_2(1+\frac{p^{(s)^{(k)}}_n h^{(s)}_n}{c_nU^{(s)}_0})} &\nonumber\\
		&-\frac{p_n^{(s)^{(k)}}H_n^{(s)}h_n^{(s)}}{c_n\bigg(\mbox{log}_2(1+\frac{p^{(s)^{(k)}}_n h^{(s)}_n}{c_nU^{(s)}_0})\bigg)^2(c_nU_0^{(s)}+p_n^{(s)^{(k)}}h_n^{(s)})\mbox{ln}2}. \nonumber
	\end{align}

By substituting (\ref{SCA}) into (\ref{P2-a}), we obtain the following approximate convex version of problem (P2) in the $k$-th iteration as problem (P2.1):
\begin{align}
	\text{(P2.1)}:& \min_{\{p_n^{(s)},f_n^{(s)}\}} \thinspace \theta(\{a_n^{(s)},\lambda_n^{(s)}\}) \nonumber\\
	&\mbox{s.t.}\quad
    \sum_{s=0}^{S}\bigg(\sum_{n=1}^{N} a^{(s)}_n(1-\lambda_n^{(s)})\big(\kappa_n\varpi_n (f_n^{(s)})^2 \frac{Z_n^{(s)} e_n}{q_n} &\nonumber\\
    &\quad\quad~ +\xi^{(k)}(p_n^{(s)})\big) + \hat{p}\max_{n\in\mathcal{N}}\big(\frac{H_n^{(s)}}{\hat{r}_n^{(s)}}\big)\bigg)\leq E_0, \! & \\
	&\quad\quad~ (\ref{P2-b}), (\ref{P1-d}), (\ref{P1-e}).\nonumber
\end{align}

\subsubsection{Optimization of Pruning Ratio}
Then, under fixed $\{a_n^{(s)}\}$, $\{p_n^{(s)}\}$, and $\{f_n^{(s)}\}$, the pruning ratio optimization problem is formulated as
\begin{subequations}\label{p3}
	\begin{align}
	 \text{(P3)}:& \min_{\{\lambda_n^{(s)}\}} \quad \theta(\{a_n^{(s)},\lambda_n^{(s)}\}) \nonumber\\
      &~~\mbox{s.t.}\quad~~ (\ref{P1-a}) \sim (\ref{P1-c}).\nonumber
	\end{align}
\end{subequations}
Problem (P3) is a linear programming (LP) problem which is solved by standard convex optimization tools, such as CVX \cite{Grant2016CVX}.

\subsubsection{Optimization of Client Selection}
Finally, given fixed computation frequency $\{f_n^{(s)}\}$, transmit power $\{p_n^{(s)}\}$, and pruning ratio $\{\lambda_n^{(s)}\}$, we optimize the client selection indicator $\{a_n^{(s)}\}$. And thus the problem is reformulated as
\begin{subequations}\label{p4}
	\begin{align}
	 \text{(P4)}:& \min_{\{a_n^{(s)}\}} \quad
      \sum_{s=0}^S\frac{\gamma_1\vert\sum_{n=1}^{N}a_n^{(s)}\phi_n\vert^2+\gamma_2\sum_{n=1}^{N}a_n^{(s)}\lambda_n^{(s)}}{\sum_{n=1}^{N}a_n^{(s)}} \nonumber\\
      &~~~~~~~~~~+\alpha+\beta\sum_{s=0}^S\frac{1}{\sum_{n=1}^{N}a_n^{(s)}}\nonumber\\
      &\quad\mbox{s.t.}\quad~~
	   (\ref{P1-a}), (\ref{P1-b}), (\ref{P1-f}). & \nonumber
	\end{align}
\end{subequations}
Although constraints (\ref{P1-a}), (\ref{P1-b}), and (\ref{P1-f}) are convex w.r.t. $\{a_n^{(s)}\}$, the objective function involves coupled fractional and quadratic terms, making Problem (P3) still non-convex. To deal with the non-convexity, we define an auxiliary variable $\mu^{(s)}$, which satisfies $\big[\gamma_1\vert\sum_{n=1}^{N}a_n^{(s)}\phi_n\vert^2+\gamma_2\sum_{n=1}^{N}a_n^{(s)}\lambda_n^{(s)}\big]\leq\mu^{(s)}$. Then, problem (P3) is reformulated as
\begin{subequations}\label{p5}
	\begin{align}
	\text{(P5)}:& \min_{\{a_n^{(s)},\mu^{(s)}\}} \alpha+\beta\sum_{s=0}^{S}\frac{1}{\sum_{n=1}^{N}a_n^{(s)}}+\sum_{s=0}^{S}\frac{\mu^{(s)}}{\sum_{n=1}^{N}a_n^{(s)}} \nonumber\\
	&\quad~~\mbox{s.t.}\quad~
    \gamma_1\left\vert\sum_{n=1}^{N}a_n^{(s)}\phi_n\right\vert^2+\gamma_2\sum_{n=1}^{N}a_n^{(s)}\lambda_n^{(s)}\leq\mu^{(s)}, &\label{P3.1-a}\\
		&\quad\quad~~~~~~~ (\ref{P1-a}), (\ref{P1-b}), (\ref{P1-f}). & \nonumber
	\end{align}
\end{subequations}
Problem (P5) is still non-convex due to the coupling between $a_n^{(s)}$ and $\mu^{(s)}$ in the objective function. This subproblem is solved by iteratively optimizing $\{a_n^{(s)}\}$ and $\{\mu^{(s)}\}$ until convergence, where one variable is optimized while the other is fixed in each iteration.

\begin{algorithm}[t]
    \caption{Proposed Algorithm for Solving Problem (P1)}
    \label{alg1}
    \begin{algorithmic}[1]
    \State Initialize the auxiliary variable $\{\mu^{(s),(0)}\}$ and optimization variables $\{a_n^{(s),(0)},\lambda_n^{(s),(0)},p_n^{(s),(0)},f_n^{(s),(0)}\}$.
        \For{$o=1:O$}
            \State Initialize $\{p_n^{(s),(o-1)^{(0)}},f_n^{(s),(o-1)^{(0)}}\}=\{p_n^{(s),(o-1)},f_n^{(s),(o-1)}\}$; Set $k=1$.
            \Repeat
                \State Solve the problem (P2.1) under the local points $\{p_n^{(s),(o-1)^{(k-1)}},f_n^{(s),(o-1)^{(k-1)}},a_n^{(s),(o-1)},\lambda_n^{(s),(o-1)}\}$ to obtain the solution $\{p_n^{(s),(o)^{(k)}},f_n^{(s),(o)^{(k)}}\}$.
                \State Set $k=k+1$.
            \Until{the decrease of the objective value is below a predefined threshold}
            \State Update $\{p_n^{(s),(o)},f_n^{(s),(o)}\}=\{p_n^{(s),(o)^{(k)}},f_n^{(s),(o)^{(k)}}\}$.
            \State Solve problem (P3) to obtain $\{\lambda_n^{(s),(o)}\}$ under given $\{a_n^{(s),(o-1)},p_n^{(s),(o)},f_n^{(s),(o)}\}$.
            \State Solve problem (P5) to obtain $\{a_n^{(s),(o)}\}$ under given $\{\lambda_n^{(s),(o)},p_n^{(s),(o)},f_n^{(s),(o)}\}$.
            \State Obtain the final solution leading to non-increasing objective value.
        \EndFor
        \State \textbf{Output} final solution $\{a_n^{(s)*},p_n^{(s)*},f_n^{(s)*},\lambda_n^{(s)*}\}$ with the minimum objective value of problem (P1).
    \end{algorithmic}
\end{algorithm}

\subsubsection{Overall algorithm}
By alternately solving the subproblems (P2) $\sim$ (P5), we can obtain a suboptimal solution to problem (P1); via iteratively optimizing the pruning ratios, client selection, as well as the communication and computation resources, we thus obtain a series of solutions $\{a_n^{(s)*},\lambda_n^{(s)*},p_n^{(s)*},f_n^{(s)*}\}$, which lead to non-increasing objective values. The proposed algorithm for solving problem (P1) is summarized in Algorithm \ref{alg1}. Since the objective function of problem (P1) monotonically decreases with each iteration and is lower bounded, the convergence of Algorithm \ref{alg1} is guaranteed.

\begin{figure}[h]
	\centering
	\includegraphics[width=80mm]{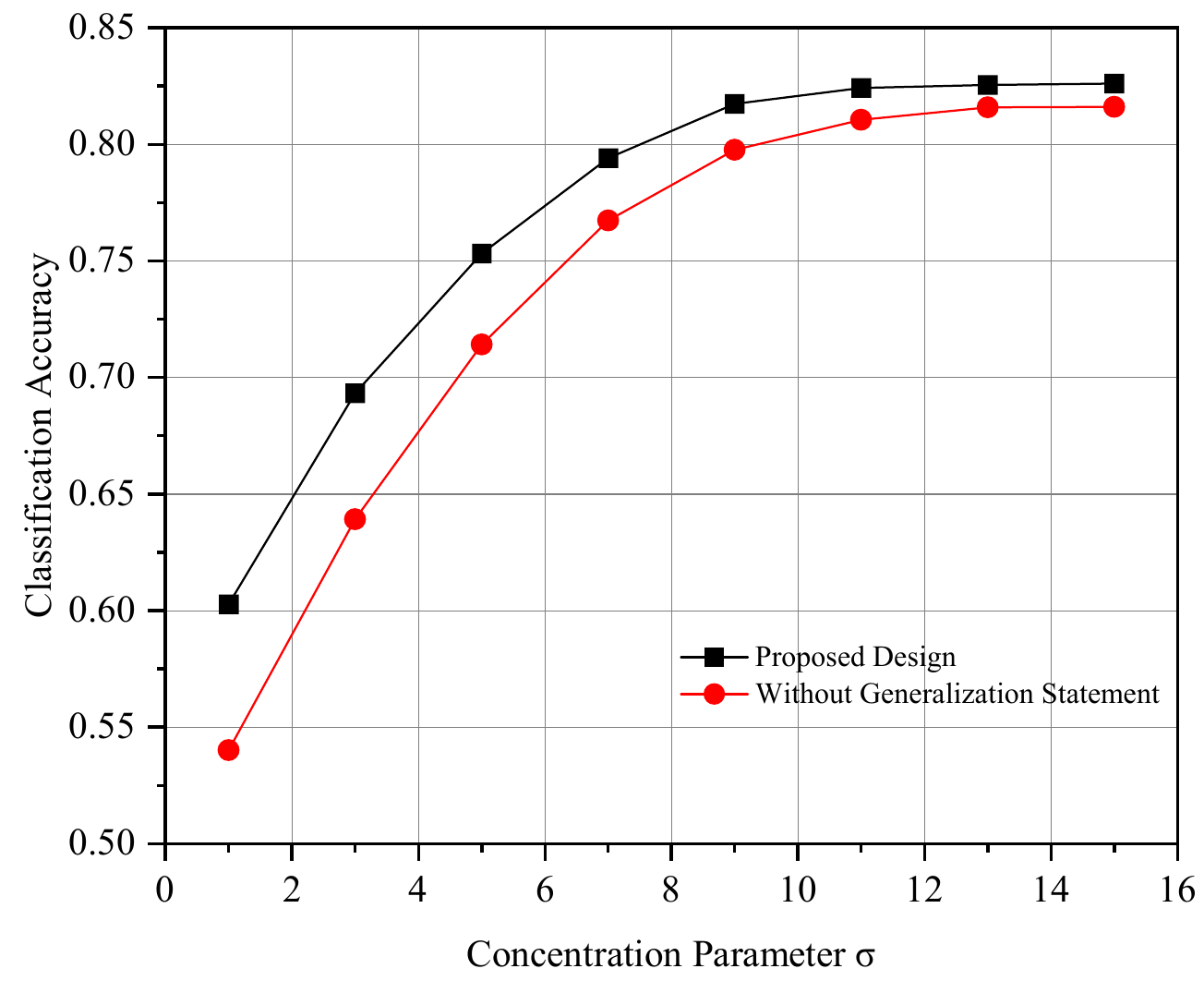}
	\caption{Classification accuracy of ResNet-110 versus Dirichlet parameter $\sigma$ with/without generalization statement under $T_0{=}3600$, and $E_0{=}7100$.}
	\label{gen}
\end{figure}

\begin{figure*}[t]
    \centering
    \subfigure[MNIST]{
        \includegraphics[width=0.44\linewidth]{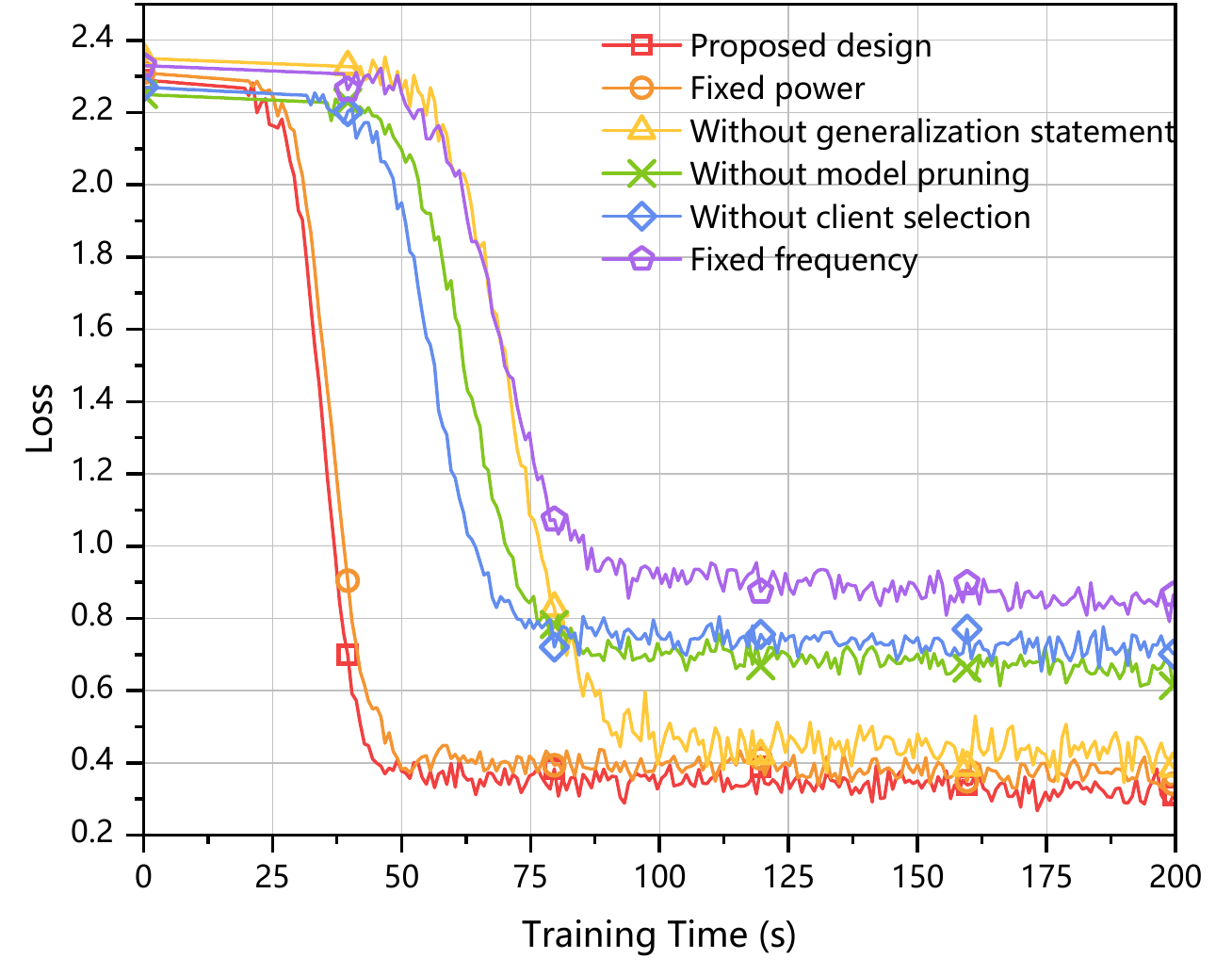}
        \label{con_t_le}}
    \subfigure[CIFAR-10]{
        \includegraphics[width=0.44\linewidth]{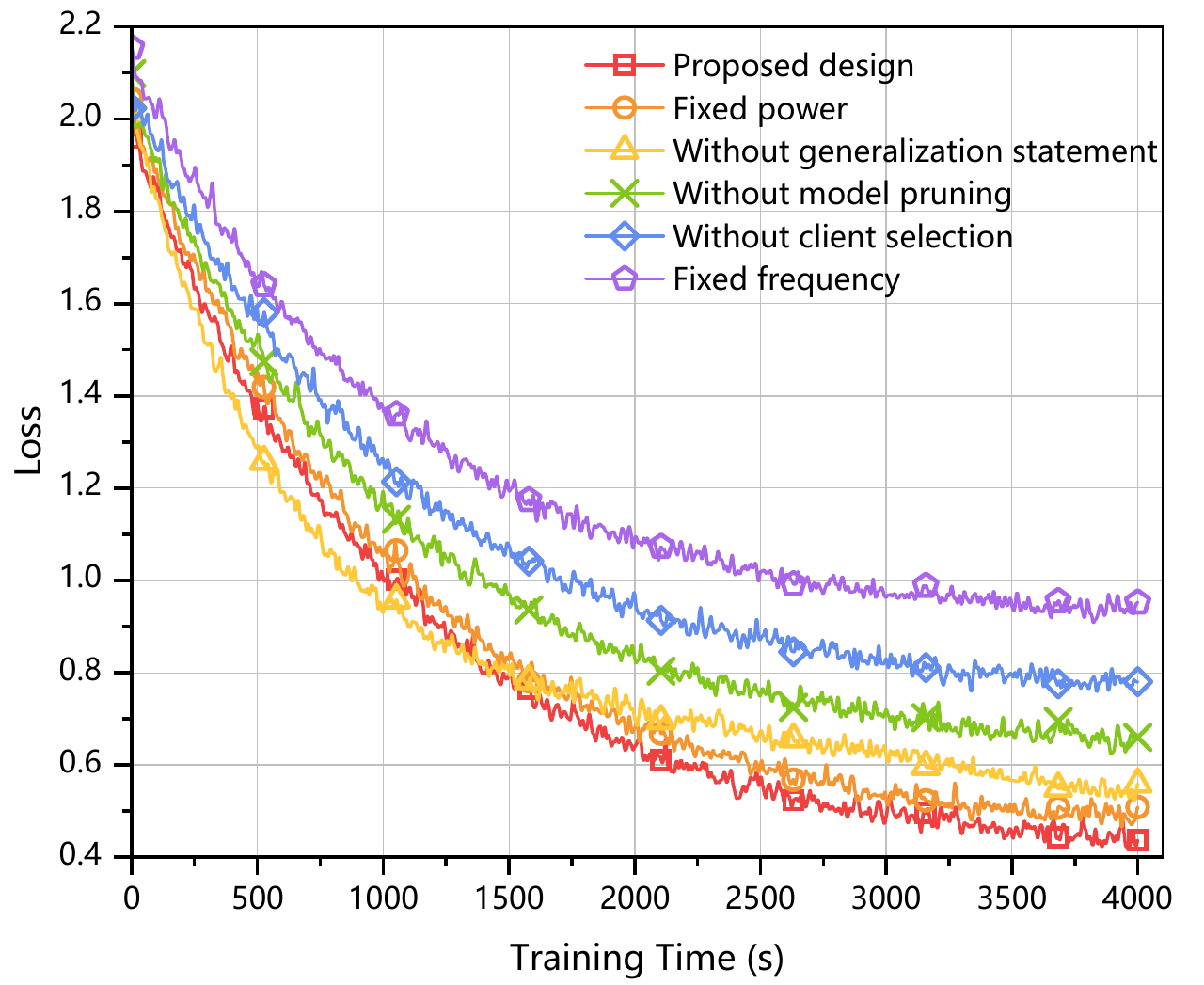}
        \label{con_t_rs}}
    \caption{Convergence behavior in terms of the training loss over the overall training time. (a) LeNet on MNIST under $E_0=250$~J; (b) ResNet-110 on CIFAR-10 under $E_0=7100$~J.}
    \label{fig:con_de}
\end{figure*}

\begin{figure*}[t]
    \centering
    \subfigure[MNIST]{
        \includegraphics[width=0.44\linewidth]{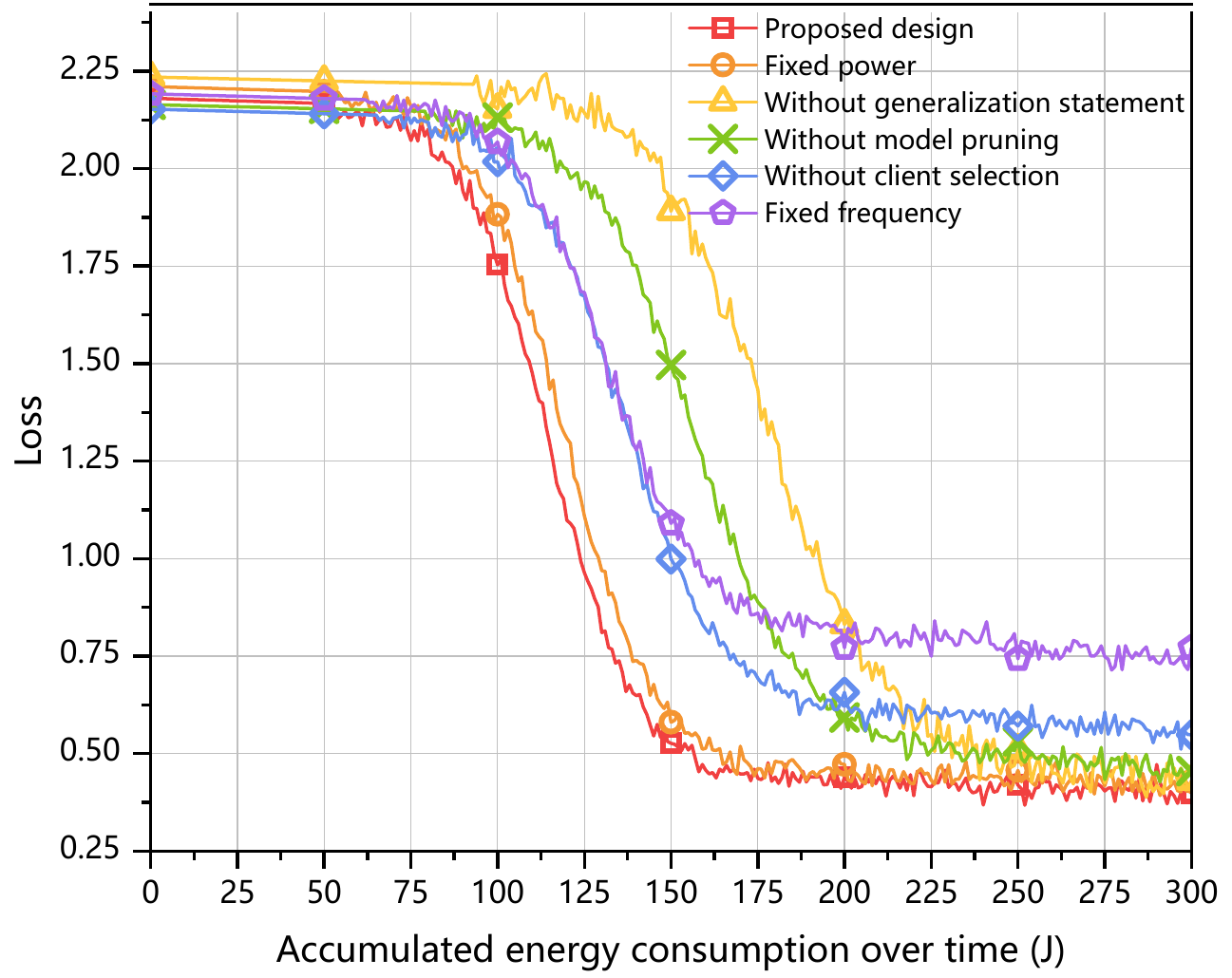}
        \label{con_e_le}}
    \subfigure[CIFAR-10]{
        \includegraphics[width=0.44\linewidth]{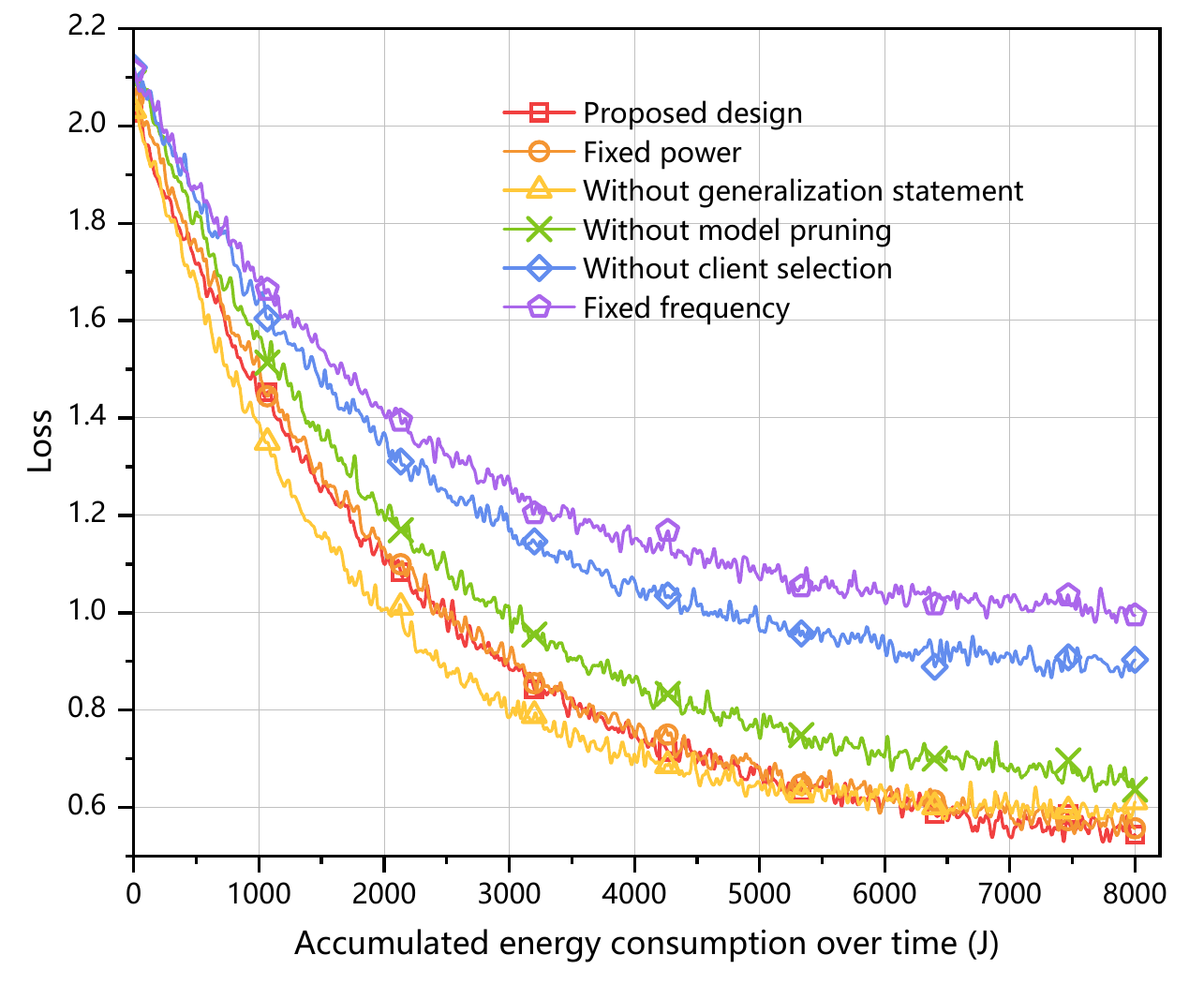}
        \label{con_e_rs}}
    \caption{Convergence behavior in terms of the training loss w.r.t. the accumulated overall system energy consumption over time. (a) LeNet on MNIST under $T_0=150$~s; (b) ResNet-110 on CIFAR-10 under $T_0=3600$~s.}
    \label{fig:con_en}
\end{figure*}

\begin{figure*}[t]
    \centering
    \subfigure[MNIST]{
        \includegraphics[width=0.44\linewidth]{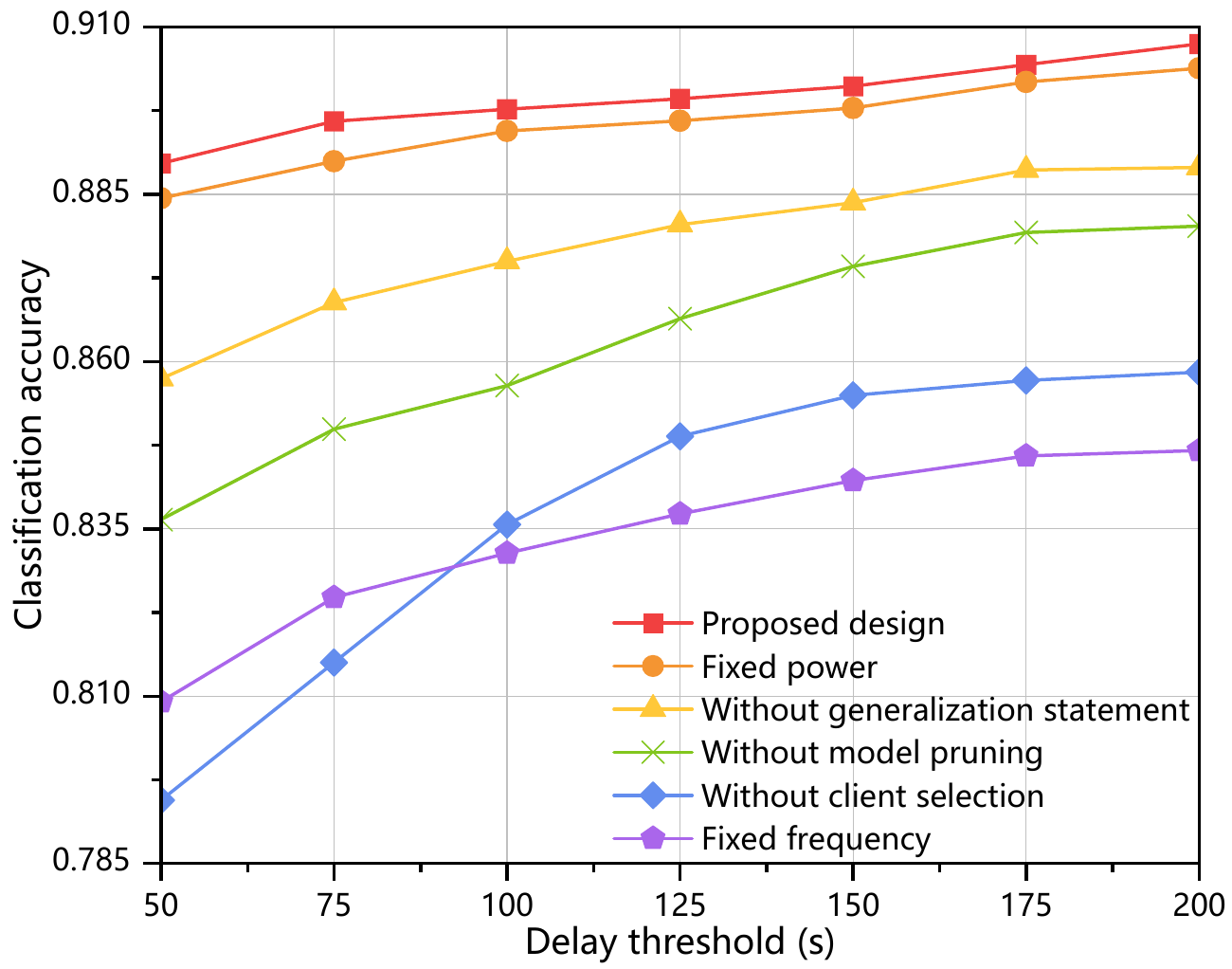}
        \label{3}}
    \subfigure[CIFAR-10]{
        \includegraphics[width=0.44\linewidth]{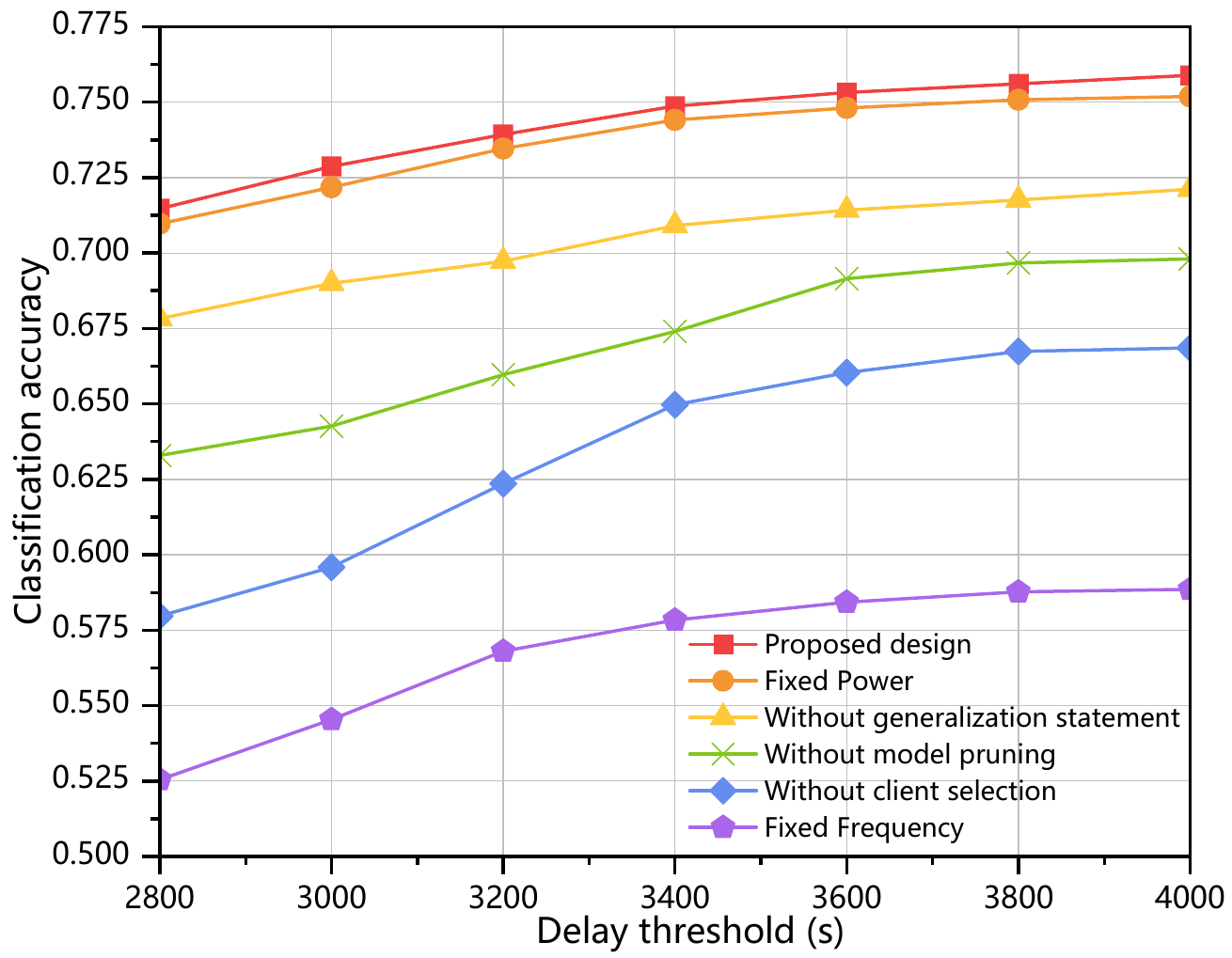}
        \label{4}}
    \caption{Classification accuracy w.r.t. different system delay thresholds (a) LeNet on MINST under  $E_0=250~\text{J}$; (b) ResNet-110 on CIFAR-10 under  $E_0=7100~\text{J}$.}
    \label{fig:acc_de}
\end{figure*}

\begin{figure*}[t]
    \centering
    \subfigure[MNIST]{
        \includegraphics[width=0.44\linewidth]{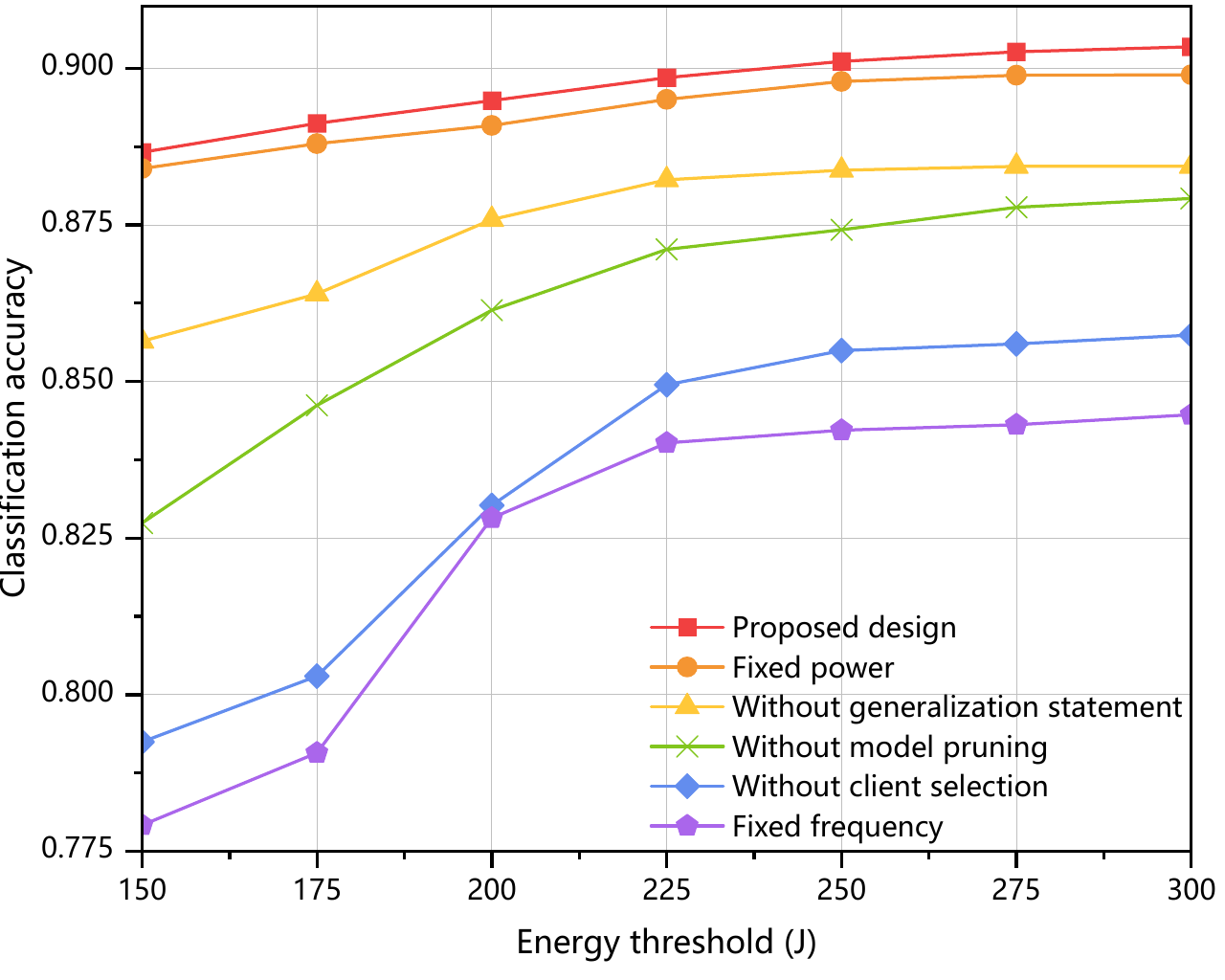}
        \label{1}}
    \subfigure[CIFAR-10]{
        \includegraphics[width=0.44\linewidth]{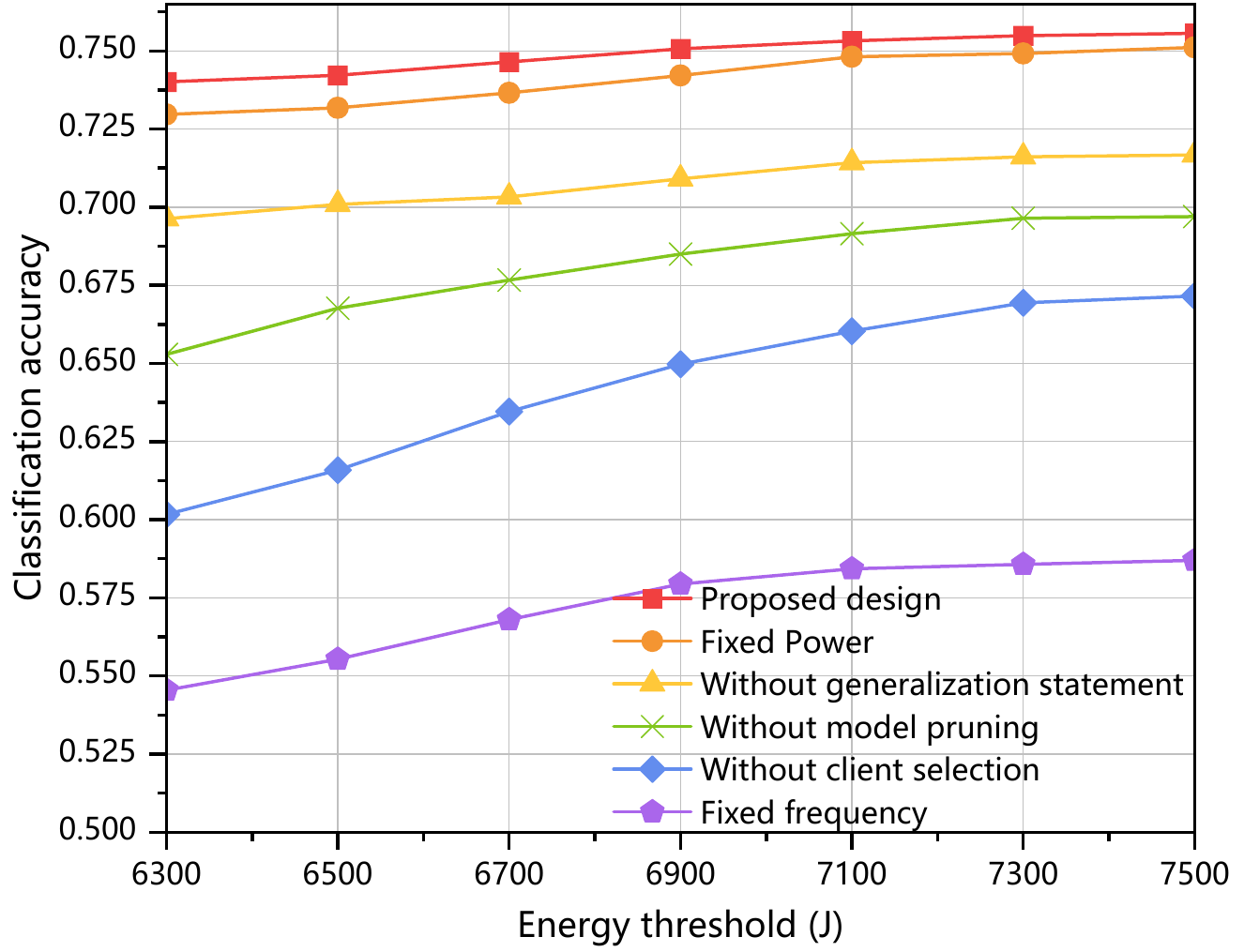}
        \label{2}}
    \caption{Classification accuracy w.r.t. different system energy thresholds (a) LeNet on MINST  under  $T_0=150~\text{s}$; (b) ResNet-110 on CIFAR-10 under  $T_0=3600~\text{s}$.}
    \label{fig:acc_en}
\end{figure*}

\section{Numerical Results}\label{sec:experiments}
This section presents numerical results to evaluate the performance of our proposed system. We set the number of clients $N=10$, learning rate $\eta=0.01$, server transmit power $\hat{p}=0.5$~W, and maximum client transmit power $p_n^{\max}=0.5$~W, $\forall n\in\mathcal{N}$. Channel coefficients are modeled as IID Rayleigh fading with an average path loss of $10^{-5}$, and remain constant during all rounds. Parameter settings are summarized in Table~\ref{tab:exp_setup}. To simulate statistical heterogeneity in distributed learning, we employ the Dirichlet-based partitioning strategy, which splits non-IID data by sampling label proportions for clients from a Dirichlet distribution $p_{n,z}\sim \text{Dirichlet}(\sigma)$, where the concentration parameter $\sigma$ controls data heterogeneity.

We evaluate LeNet~\cite{LeCun1998} on the MNIST~\cite{LeCun1998MNIST} dataset, followed by experiments with ResNet-110~\cite{He2016} on CIFAR-10~\cite{Krizhevsky2009}. MNIST contains 60,000 training and 10,000 testing grayscale images (28$\times$28), while CIFAR-10 consists of 50,000 training and 10,000 testing RGB images (32$\times$32$\times$3) from 10 object categories. All experiments are implemented using PyTorch and conducted on NVIDIA RTX 3090 GPUs.

For comparison, we consider the following benchmark schemes.
\begin{itemize}
    \item \textbf{Fixed pruning:}~The model is trained without pruning, i.e., $\lambda_n^{(s)}=0,~\forall~n,s$, to evaluate the effect of pruning ratio optimization.
    \item \textbf{Fixed selection:}~All clients participate in each training round by setting $a_n^{(s)}=1,~\forall~n,s$, serving as the baseline without client selection.
    \item \textbf{Without generalization statement:}~This scheme follows the conventional convergence analysis in~\cite{Liu20222}, without generalization statement based optimization.
    \item \textbf{Fixed power design:}~Each client transmits with a constant power $p_n^{(s)}=0.5$~W$,~\forall n,s$, to examine the benefit of adaptive power control.
    \item \textbf{Fixed clock frequency design:}~We set the computation frequency of each client as $f_n^{(s)}=f^{\max},~\forall n,s$, for comparison with adaptive computation design.
\end{itemize}

Figs. \ref{fig:data_dis}\subref{5} and \ref{fig:data_dis}\subref{8} illustrate the impact of client data heterogeneity on label distribution and generalization statement under different Dirichlet parameters $\sigma$. As $\sigma$ increases, the label distributions among clients become more balanced, and the variation in generalization performance decreases. Fig. \ref{gen} shows that incorporating the generalization statement improves classification accuracy across different levels of data heterogeneity, consistently outperforming the baseline without it. To evaluate the generalization ability, we conduct experiments on $\sigma=5$ to simulate data heterogeneity \cite{Guo2025}.

Figs.~\ref{fig:con_de}\subref{con_t_le} and \ref{fig:con_de}\subref{con_t_rs} demonstrate the convergence performance in terms of training loss w.r.t. the training time. It is observed that, within the permissible training delay threshold, the proposed system consistently attains the lowest loss values across all schemes. While certain baseline methods for model ResNet-110 show a faster initial decrease in loss, they finally converge to a higher loss compared to the proposed design. This demonstrates the effectiveness of the proposed system in dynamically adjusting client participation and compression levels based on local data quality, particularly under non-IID conditions. 

Figs. \ref{fig:con_en}\subref{con_e_le} and \ref{fig:con_en}\subref{con_e_rs} demonstrate the convergence performance of training loss w.r.t the accumulated system energy consumption. It is observed that under a permissible energy constraint, the proposed framework achieves superior training performance, as evidenced by lower final loss values. Although some baselines exhibit faster loss reduction in the early training stages, their final performance remains suboptimal. This demonstrates that, by efficiently coordinating model selection, pruning, and resource allocation under constrained energy resources, the proposed design leads to more favorable final convergence behavior.

Figs. \ref{fig:acc_de}\subref{3} and \ref{fig:acc_de}\subref{4} show the classification accuracy under different system delay thresholds. It is observed that the proposed design achieves higher accuracy than baseline schemes. As the energy consumption threshold increases, the performance improvement of all baselines gradually saturates, and latency emerges as the dominant system constraint. Notably, the proposed design significantly outperforms the design without generalization statement. This demonstrates that, in joint resource optimization, generalization statements enable effective selection of both clients and pruning levels, thereby yielding higher test accuracy and superior model generalization. 

Figs. \ref{fig:acc_en}\subref{1} and \ref{fig:acc_en}\subref{2} show the classification accuracy under energy consumption thresholds. It is observed that under tight resource limitations, the proposed design exhibits a considerable accuracy advantage, demonstrating the effectiveness of joint resource optimization. Furthermore, compared with the system without generalization statements, the proposed design achieves higher accuracy, indicating the advantage of incorporating generalization analysis into system optimization.

\section{Conclusions}\label{sec:conclusion}
This paper presented a parameter-efficient FEEL framework optimized through generalization analysis to improve both model generalization and resource utilization in resource-constrained edge deployment. We first established an information-theoretic generalization statement to quantify the divergence between local training and testing function losses, and analyzed the bound of average squared gradient norm.  Then, we designed an efficient algorithm to solve the joint optimization problem of minimizing the above bound over client selection, model pruning, and resource allocation. Finally, numerical results demonstrated that under energy and delay constraints, the proposed design achieves better convergence and generalization performance. Future work will further explore adaptive local training data sampling strategies under dynamic environments to enhance robustness against temporal data shifts. In addition, extending the proposed framework to large-scale model fine-tuning is another important direction toward realizing more general and scalable edge AI.

\appendices
\section{Proof of Lemma \ref{lem1}}
\label{appendix:lemma1}
To analyze generalization, we first derive the expression for the gradient discrepancy norm between training and testing sets by decomposing the loss gradient as
\begin{align}\label{37}
	\Vert\nabla&\mathcal{L}(\boldsymbol{\omega},\hat{\mathcal{D}})-\nabla\tilde{\mathcal{L}}(\boldsymbol{\omega},\tilde{\mathcal{D}})\Vert \nonumber\\
	&=\big\Vert\sum_{z\in\mathcal{D}}\nabla l(\boldsymbol{\omega},z)[\mathrm{p}(z\vert\hat{\mathcal{D}})-\mathrm{p}(z\vert\tilde{\mathcal{D}})]\big\Vert \nonumber\\
	&=\big\Vert\sum_{z\in\hat{\mathcal{D}}\cup\tilde{\mathcal{D}}}\nabla l(\boldsymbol{\omega},z)\vert \mathrm{p}(z\vert\hat{\mathcal{D}})-\mathrm{p}(z\vert\tilde{\mathcal{D}})\vert\big\Vert \nonumber\\
	&\leq\big\Vert\sum_{z\in\hat{\mathcal{D}}\cup\tilde{\mathcal{D}}}\nabla l(\boldsymbol{\omega},z)\sum_{z\in\hat{\mathcal{D}}\cup\tilde{\mathcal{D}}}\vert \mathrm{p}(z\vert \hat{\mathcal{D}})-\mathrm{p}(z\vert\tilde{\mathcal{D}})\vert\big\Vert,
\end{align}
where $l(\boldsymbol{\omega},z)$ denotes the loss function of model $\boldsymbol{\omega}$ on data sample $z$ from the dataset. According to Pinsker's inequality, we have $\vert \mathrm{p}(\hat{\mathcal{D}})-\mathrm{p}(\tilde{\mathcal{D}})\vert\leq\sqrt{2D_{KL}[\mathrm{p}(\hat{\mathcal{D}})\Vert\mathrm{p}(\tilde{\mathcal{D}})]}$, where $\mathrm{p}(\hat{\mathcal{D}})$ and $\mathrm{p}(\tilde{\mathcal{D}})$ denote the training and test data distributions, respectively, and $D_{KL}[\mathrm{p}(\hat{\mathcal{D}})\Vert\mathrm{p}(\tilde{\mathcal{D}})]$ denotes the Kullback–Leibler (KL) divergence of the above distributions. Accordingly, inequality~(\ref{37}) is established as
\begin{align}\label{A-1}
	\Vert\nabla&\mathcal{L}(\boldsymbol{\omega},\hat{\mathcal{D}})-\nabla\tilde{\mathcal{L}}(\boldsymbol{\omega},\tilde{\mathcal{D}})\Vert \nonumber\\
	&\leq\big\Vert\sum_{z\in\hat{\mathcal{D}}\cup\tilde{\mathcal{D}}}\nabla l(\boldsymbol{\omega},z)\big\Vert\sqrt{2D_{KL}[\mathrm{p}(\hat{\mathcal{D}})\Vert\mathrm{p}(\tilde{\mathcal{D}})]} \nonumber\\
	&\leq\big\Vert\sum_{z\in\hat{\mathcal{D}}}\nabla l(\boldsymbol{\omega},z)\big\Vert\sqrt{2D_{KL}[\mathrm{p}(\hat{\mathcal{D}})\Vert\mathrm{p}(\tilde{\mathcal{D}})]} \nonumber\\
	&+\big\Vert\sum_{z\in\tilde{\mathcal{D}}}\nabla l(\boldsymbol{\omega},z)\big\Vert\sqrt{2D_{KL}[\mathrm{p}(\hat{\mathcal{D}})\Vert\mathrm{p}(\tilde{\mathcal{D}})]} \nonumber\\
	&\leq\hat{D}\Vert\nabla\mathcal{L}(\boldsymbol{\omega},\hat{\mathcal{D}})\Vert\sqrt{2D_{KL}[\mathrm{p}(\hat{\mathcal{D}})\Vert\mathrm{p}(\tilde{\mathcal{D}})]} \nonumber\\
    &+\tilde{D}\Vert\nabla\tilde{\mathcal{L}}(\boldsymbol{\omega},\tilde{\mathcal{D}})\Vert\sqrt{2D_{KL}[\mathrm{p}(\hat{\mathcal{D}})\Vert\mathrm{p}(\tilde{\mathcal{D}})]}.
\end{align}
Then, we apply the reverse triangle inequality $\vert\Vert\nabla\mathcal{L}(\boldsymbol{\omega},\hat{\mathcal{D}})\Vert-\Vert\nabla\tilde{\mathcal{L}}(\boldsymbol{\omega},\tilde{\mathcal{D}})\Vert\vert\leq\Vert\nabla\mathcal{L}(\boldsymbol{\omega},\hat{\mathcal{D}})-\nabla\tilde{\mathcal{L}}(\boldsymbol{\omega},\tilde{\mathcal{D}})\Vert$ to~(\ref{A-1}), which yields
\begin{align}\label{A-2}
	&\frac{1-\tilde{D}\sqrt{2D_{KL}[\mathrm{p}(\hat{\mathcal{D}})\Vert\mathrm{p}(\tilde{\mathcal{D}})]}}{1+\hat{D}\sqrt{2D_{KL}[\mathrm{p}(\hat{\mathcal{D}})\Vert\mathrm{p}(\tilde{\mathcal{D}})]}}\Vert\nabla\tilde{\mathcal{L}}(\boldsymbol{\omega},\tilde{\mathcal{D}})\Vert\leq\Vert\nabla\mathcal{L}(\boldsymbol{\omega},\hat{\mathcal{D}})\Vert.
\end{align}

Next, to analyze the above inequality, we consider two cases based on the sign of $\big(1 - \tilde{D}\sqrt{2D_{\mathrm{KL}}[\mathrm{p}(\hat{\mathcal{D}}) \Vert\mathrm{p}(\tilde{\mathcal{D}})]}\big)$. Firstly, when $1 - \tilde{D}\sqrt{2D_{\mathrm{KL}}[\mathrm{p}(\hat{\mathcal{D}}) \Vert\mathrm{p}(\tilde{\mathcal{D}})]} > 0$, we substitute (\ref{A-2}) into (\ref{A-1}) and obtain
\begin{align}\label{A-3}
	&\Vert\nabla\mathcal{L}(\boldsymbol{\omega},\hat{\mathcal{D}})-\nabla\tilde{\mathcal{L}}(\boldsymbol{\omega},\tilde{\mathcal{D}})\Vert \nonumber\\
	&\leq\big(\hat{D}\Vert\nabla\mathcal{L}(\boldsymbol{\omega},\hat{\mathcal{D}})\Vert+\tilde{D}\Vert\nabla\tilde{\mathcal{L}}(\boldsymbol{\omega},\tilde{\mathcal{D}})\Vert\big)\sqrt{2D_{KL}[\mathrm{p}(\hat{\mathcal{D}})\Vert\mathrm{p}(\tilde{\mathcal{D}})]} \nonumber\\
    &\leq\big(\tilde{D}\cdot\frac{1+\hat{D}\sqrt{2D_{KL}[\mathrm{p}(\hat{\mathcal{D}})\Vert\mathrm{p}(\tilde{\mathcal{D}})]}}{1-\tilde{D}\sqrt{2D_{KL}[\mathrm{p}(\hat{\mathcal{D}})\Vert\mathrm{p}(\tilde{\mathcal{D}})]}} \nonumber\\
	&+\hat{D}\big)\Vert\nabla\mathcal{L}(\boldsymbol{\omega},\hat{\mathcal{D}})\Vert\sqrt{2D_{KL}[\mathrm{p}(\hat{\mathcal{D}})\Vert\mathrm{p}(\tilde{\mathcal{D}})]} \nonumber\\
	&=\bigg\vert\frac{(\hat{D}+\tilde{D}_n)\sqrt{2D_{KL}[\mathrm{p}(\hat{\mathcal{D}})\Vert\mathrm{p}(\tilde{\mathcal{D}})]}}{1-\tilde{D}\sqrt{2D_{KL}[\mathrm{p}(\hat{\mathcal{D}})\Vert\mathrm{p}(\tilde{\mathcal{D}})]}}\bigg\vert\Vert\nabla\mathcal{L}(\boldsymbol{\omega},\hat{\mathcal{D}})\Vert \nonumber\\
    &=\phi^{\prime}\Vert\nabla\mathcal{L}(\boldsymbol{\omega},\hat{\mathcal{D}})\Vert,
\end{align}
where we define the preliminary generalization gap term as $\phi^{\prime}$. Next, when $\big(1-\tilde{D}\sqrt{2D_{KL}[\mathrm{p}(\hat{\mathcal{D}})\Vert\mathrm{p}(\tilde{\mathcal{D}})]}\big)\leq 0$, it follows that $\phi^{\prime}\geq 1$. Consequently, we have
$\vert\mathrm{p}(z\vert\hat{\mathcal{D}})-\mathrm{p}(z\vert\tilde{\mathcal{D}})\vert\leq1\leq\phi^{\prime}$,
which leads to
\begin{align}
	&\Vert\nabla\mathcal{L}(\boldsymbol{\omega},\hat{\mathcal{D}})-\nabla\tilde{\mathcal{L}}(\boldsymbol{\omega},\tilde{\mathcal{D}})\Vert \nonumber\\
    &=\big\Vert\sum_{z\in\hat{\mathcal{D}}\cup\tilde{\mathcal{D}}}\nabla l(\boldsymbol{\omega},z)[\mathrm{p}(z\vert\hat{\mathcal{D}})-\mathrm{p}(z\vert\tilde{\mathcal{D}})]\big\Vert \nonumber\\
	&\leq\big\Vert\sum_{z\in\hat{\mathcal{D}}\cup\tilde{\mathcal{D}}}\nabla l(\boldsymbol{\omega},z)\phi^{\prime}\mathrm{p}(z\vert\hat{\mathcal{D}})\frac{1}{\mathrm{p}(z\vert\hat{\mathcal{D}})}\big\Vert \nonumber\\
	&\leq\phi^{\prime}\frac{1}{\mathrm{p}^{\prime}(z\vert\hat{\mathcal{D}})}\Vert\nabla\mathcal{L}(\boldsymbol{\omega},\hat{\mathcal{D}})\Vert,
\end{align}
where $\mathrm{p}^{\prime}(z\vert\hat{\mathcal{D}})$ denotes the probability of the least frequent element in the distribution. KL divergence is decomposed as
\begin{align}
	\mathrm{D}_{\mathrm{KL}}[\mathrm{p}(z\vert\hat{\mathcal{D}})\Vert\mathrm{p}&(z\vert\tilde{\mathcal{D}})]=\mathrm{H}\big(\mathrm{p}(z\vert\hat{\mathcal{D}}),\mathrm{p}(z\vert\tilde{\mathcal{D}})\big)-\mathrm{H}\big(\mathrm{p}(z\vert\hat{\mathcal{D}})\big) \nonumber\\
	&=\mathrm{H}(\mathrm{p}(z\vert\tilde{\mathcal{D}}))-\big[\mathrm{H}(\mathrm{p}(z\vert\hat{\mathcal{D}})) \nonumber\\
	&+\mathrm{H}(\mathrm{p}(z\vert\tilde{\mathcal{D}}))-\mathrm{H}\big(\mathrm{p}(z\vert\hat{\mathcal{D}}),\mathrm{p}(z\vert\tilde{\mathcal{D}})\big)\big] \nonumber\\
	&=\mathrm{H}(\mathrm{p}(z\vert\tilde{\mathcal{D}}))-\mathrm{I}\big(\mathrm{p}(z\vert\hat{\mathcal{D}}),\mathrm{p}(z\vert\tilde{\mathcal{D}})\big).
\end{align}

Finally, the norm of the difference between the gradients computed on the training and test datasets is bounded as 
\begin{align}\label{A-4}
	\Vert\nabla\mathcal{L}(\boldsymbol{\omega},\hat{\mathcal{D}})-\nabla\tilde{\mathcal{L}}(\boldsymbol{\omega},\tilde{\mathcal{D}})\Vert\leq\phi\Vert\nabla\mathcal{L}(\boldsymbol{\omega},\hat{\mathcal{D}})\Vert,
\end{align}
where the generalization statement is defined as
\begin{align}
    \phi=\frac{(\hat{D}+\tilde{D})}{\mathrm{p}^{'}(z\vert\hat{\mathcal{D}})}
    \cdot \nonumber\bigg\vert\frac{\sqrt{2\big(\mathrm{H}(\mathrm{p}(z\vert\tilde{\mathcal{D}}))-\mathrm{I}(\mathrm{p}(z\vert\hat{\mathcal{D}}),\mathrm{p}(z\vert\tilde{\mathcal{D}}))\big)}}{1-\tilde{D}\sqrt{2\big(\mathrm{H}(\mathrm{p}(z\vert\tilde{\mathcal{D}}))-\mathrm{I}(\mathrm{p}(z\vert\hat{\mathcal{D}}),\mathrm{p}(z\vert\tilde{\mathcal{D}}))\big)}}\bigg\vert.
\end{align}
Here, (\ref{A-4}) bounds the gradient gap between training and test sets via gradients. $\phi$ reflects distributional shift and scale variation. Smaller values indicate better generalization, while larger ones imply performance degradation on unseen data.

\section{Proof of Proposition \ref{lem2}}
\label{appendix:proposition1}
For notational simplicity, we use $\mathcal{L}(\boldsymbol{\omega})$ to denote $\mathcal{L}(\boldsymbol{\omega},\hat{\mathcal{D})}$ and $\tilde{\mathcal{L}}(\boldsymbol{\omega})$ to denote $\tilde{\mathcal{L}}(\boldsymbol{\omega},\tilde{\mathcal{D}})$ in the following derivations. To start with, by leveraging gradient descent and Taylor expansion, we decompose the loss function as 
\begin{align}
	\mathcal{L}&(\boldsymbol{\omega}^{(s+1)})=\mathcal{L}\big(\boldsymbol{\omega}^{(s)}-\eta G(\tilde{\boldsymbol{\omega}}^{(s)})\big) \nonumber\\
	&=\mathcal{L}(\boldsymbol{\omega}^{(s)})-\eta G(\tilde{\boldsymbol{\omega}}^{(s)})^{\mathrm{T}}\nabla\mathcal{L}(\boldsymbol{\omega}^{(s)})+\mathcal{O}\big(\eta G(\tilde{\boldsymbol{\omega}}^{(s)})\big).
\end{align}
The generalization gap between two iterations is
\begin{align}\label{45}
	\varphi&^{(s+1)}-\varphi^{(s)} \nonumber \\
	&=\mathbb{E}[[\mathcal{L}(\boldsymbol{\omega}^{(s+1)})-\tilde{\mathcal{L}}(\boldsymbol{\omega}^{(s+1)})]-[\mathcal{L}(\boldsymbol{\omega}^{(s)})-\tilde{\mathcal{L}}(\boldsymbol{\omega}^{(s)})]] \nonumber\\
	&=\mathbb{E}[[\mathcal{L}(\boldsymbol{\omega}^{(s+1)})-\mathcal{L}(\boldsymbol{\omega}^{(s)})]-[\tilde{\mathcal{L}}(\boldsymbol{\omega}^{(s+1)})-\tilde{\mathcal{L}}(\boldsymbol{\omega}^{(s)})]] \nonumber\\
	&=\mathbb{E}\big[\eta G(\tilde{\boldsymbol{\omega}}^{(s)})^{\mathrm{T}}[\nabla\tilde{\mathcal{L}}(\boldsymbol{\omega}^{(s)})-\nabla\mathcal{L}(\boldsymbol{\omega}^{(s)})]\big].
\end{align}
Then, using the Young's inequality $a^{\mathrm{T}}b\leq\frac{1}{2}(\Vert a\Vert^2+\Vert b\Vert^2)$, we bound (\ref{45}) as
\begin{align}\label{46}
    &\varphi^{(s+1)}-\varphi^{(s)} \nonumber\\
    &=\mathbb{E}\big[\eta G(\tilde{\boldsymbol{\omega}}^{(s)})^{\mathrm{T}}[\nabla\tilde{\mathcal{L}}(\boldsymbol{\omega}^{(s)})-\nabla\mathcal{L}(\boldsymbol{\omega}^{(s)})]\big] \nonumber\\
    &\leq\mathbb{E}\big[\frac{1}{2}(\eta^2\Vert G(\tilde{\boldsymbol{\omega}}^{(s)})\Vert^2+\Vert\nabla\tilde{\mathcal{L}}(\boldsymbol{\omega}^{(s)})-\nabla\mathcal{L}(\boldsymbol{\omega}^{(s)})\Vert^2)\big].
\end{align}
Next, to further bound (\ref{46}), we make the following analysis. In FEEL, each client is restricted to a limited and biased local subset that poorly represents the global distribution, leading to larger generalization statements. Consequently, the generalization statement derived from the global dataset is considerably lower than that of the selected clients. Based on this observation and Lemma \ref{lem1}, we establish the bound
\begin{align}
	\Vert\nabla\mathcal{L}(\boldsymbol{\omega}^{(s)})-\nabla\tilde{\mathcal{L}}(\boldsymbol{\omega}^{(s)})\Vert&\leq\hat{\phi}\Vert\nabla\mathcal{L}(\boldsymbol{\omega}^{(s)})\Vert \nonumber\\
    &\leq\sum_{n=1}^{N}a_n\phi_n\Vert\nabla\mathcal{L}(\boldsymbol{\omega}^{(s)})\Vert,
\end{align}
where $\hat{\phi}$ denotes the generalization statement of global dataset. Accordingly, we bound the generalization statement between rounds $s$ and $(s+1)$ as
\begin{align}
	\varphi^{(s+1)}&-\varphi^{(s)}\leq\frac{1}{2}\big(\eta^2+\vert\sum_{n=1}^{N}a_n\phi_n\vert^2\big)\mathbb{E}\left\{\Vert G(\tilde{\boldsymbol{\omega}}^{(s)})\Vert^2\right\}.
\end{align}

\section{Proof of Theorem \ref{theo1}}
\label{appendix:theorem1}
To analyze the convergence property, we first decompose the loss function as
\begin{align}
	\mathcal{L}(\boldsymbol{\omega}^{(s+1)})&=\tilde{\mathcal{L}}(\boldsymbol{\omega}^{(s+1)})+\varphi^{(s+1)} \nonumber\\
	&\leq\tilde{\mathcal{L}}(\boldsymbol{\omega}^{(s)})-\eta\langle\nabla\tilde{\mathcal{L}}(\boldsymbol{\omega}^{(s)}),\nabla\tilde{\mathcal{L}}(\tilde{\boldsymbol{\omega}}^{(s)})\rangle \nonumber\\
	&+\frac{\eta^2L}{2}\mathbb{E}\left\{\Vert G(\tilde{\boldsymbol{\omega}}^{(s)})\Vert^2\right\}+\varphi^{(s+1)}.
\end{align}

We suppose that the local model trained with a fixed batch size $Z_n^{(s)}=Z,\forall n,s$. Then, the aggregated global gradient $G(\tilde{\boldsymbol{\omega}}^{(s)})$ is decomposed as
\begin{align}
	\Vert G(\tilde{\boldsymbol{\omega}}^{(s)})\Vert^2&=\Vert\frac{1}{\tilde{N}^{(s)}}\sum^N_{n=1}a_n^{(s)}g(\tilde{\boldsymbol{\omega}}_n^{(s)})\Vert^2 \nonumber\\
	&=\Vert\frac{1}{\tilde{N}^{(s)}}\sum^N_{n=1}\frac{a_n^{(s)}}{Z}\sum_{i=1}^Zg^{(i)}(\tilde{\boldsymbol{\omega}}_n^{(s)})\Vert^2 \nonumber\\
	&=\Vert \frac{1}{Z\tilde{N}^{(s)}}\sum_{n=1}^{N}\sum_{i=1}^{Z}a_n^{(s)}g^{(i)}(\tilde{\boldsymbol{\omega}}_n^{(s)})\Vert^2.
\end{align}
Then, from Assumption \ref{Assump3}, we have
\begin{align}
	\mathbb{E}[\Vert G(\tilde{\boldsymbol{\omega}}^{(s)})\Vert^2]\leq\frac{A^2}{Z\tilde{N}^{(s)}}=\frac{A^2}{\mathrm{Z}^{(s)}},
\end{align}
where $\mathrm{Z}^{(s)}=\sum_{n=1}^{N}a_n^{(s)}Z$ is the total batch size in round $s$. Accordingly, we bound the loss function as
\begin{align}\label{52}
	&\mathcal{L}(\boldsymbol{\omega}^{(s+1)})\leq\tilde{\mathcal{L}}(\boldsymbol{\omega}^{(s)})-\eta\langle\nabla\tilde{\mathcal{L}}(\boldsymbol{\omega}^{(s)}),\nabla\tilde{\mathcal{L}}(\tilde{\boldsymbol{\omega}}^{(s)})\rangle+\varphi^{(s)} \nonumber\\
	&+\frac{1}{2}\bigg(\eta^2L+\eta^2+\vert\sum_{n=1}^{N}a_n\phi_n\vert^2\bigg)\mathbb{E}\left\{\Vert G(\tilde{\boldsymbol{\omega}}^{(s)})\Vert^2\right\} \nonumber\\
	&\leq\tilde{\mathcal{L}}(\boldsymbol{\omega}^{(s)})-\eta\langle\nabla\tilde{\mathcal{L}}(\boldsymbol{\omega}^{(s)}),\nabla\tilde{\mathcal{L}}(\tilde{\boldsymbol{\omega}}^{(s)})\rangle+\varphi^{(s)} \nonumber\\
	&+\frac{A^2}{2\mathrm{Z}^{(s)}}\bigg(\eta^2L+\eta^2+\vert\sum_{n=1}^{N}a_n\phi_n\vert^2\bigg) \nonumber\\
	&\leq\tilde{\mathcal{L}}(\boldsymbol{\omega}^{(s)})+\varphi^{(s)}-\eta\langle\nabla\tilde{\mathcal{L}}(\tilde{\boldsymbol{\omega}}^{(s)}),\nabla\tilde{\mathcal{L}}(\tilde{\boldsymbol{\omega}}^{(s)})\rangle \nonumber\\
	&+\frac{A^2}{2Z}\bigg(\eta^2L+\eta^2+\vert\sum_{n=1}^{N}a_n\phi_n\vert^2\bigg)\frac{1}{\sum_{n=1}^{N}a_n^{(s)}} \nonumber\\
	&+\eta\langle\nabla\tilde{\mathcal{L}}(\tilde{\boldsymbol{\omega}}^{(s)})-\nabla\tilde{\mathcal{L}}(\boldsymbol{\omega}^{(s)}),\nabla\tilde{\mathcal{L}}(\tilde{\boldsymbol{\omega}}^{(s)})\rangle.
\end{align}
By applying Assumption \ref{Assump1} and Young’s inequality to (\ref{52}), we further obtain
\begin{align}\label{53}
	&\mathcal{L}(\boldsymbol{\omega}^{(s+1)})\leq\frac{A^2}{2Z}\bigg(\eta^2L+\eta^2+\vert\sum_{n=1}^{N}a_n\phi_n\vert^2\bigg)\frac{1}{\sum_{n=1}^{N}a_n^{(s)}} \nonumber\\
	&+\frac{\eta L^2}{2\tilde{N}^{(s)}}\sum_{n\in\tilde{\mathcal{N}}^{(s)}}\Vert\boldsymbol{\omega}_n^{(s)}-\tilde{\boldsymbol{\omega}}_n^{(s)}\Vert^2-\frac{\eta}{2}\Vert\nabla\tilde{\mathcal{L}}(\tilde{\boldsymbol{\omega}}^{(s)})\Vert^2 \nonumber\\
    &+\tilde{\mathcal{L}}(\boldsymbol{\omega}^{(s)})+\varphi^{(s)}.
\end{align}
Based on $\mathcal{L}(\boldsymbol{\omega}^{(s)})=\tilde{\mathcal{L}}(\boldsymbol{\omega}^{(s)})+\varphi^{(s)}$, we update (\ref{53}) as
\begin{align}\label{55}
	\mathbb{E}&\left\{\Vert \nabla\tilde{\mathcal{L}}(\tilde{\boldsymbol{\omega}}^{(s)})\Vert^2\right\}\leq-\frac{2}{\eta}(\mathbb{E}\left\{\mathcal{L}(\boldsymbol{\omega}^{(s+1)})\right\}-\mathbb{E}\left\{\mathcal{L}(\boldsymbol{\omega}^{(s)})\right\}) \nonumber\\
	&+\frac{\eta A^2}{Z}\bigg(\eta^2L+\eta^2+\vert\sum_{n=1}^{N}a_n\phi_n\vert^2\bigg)\frac{1}{\sum_{n=1}^{N}a_n^{(s)}}  \nonumber\\
	&+\frac{L^2}{\tilde{N}^{(s)}}\sum_{n\in\tilde{\mathcal{N}}^{(s)}}\Vert\boldsymbol{\omega}_n^{(s)}-\tilde{\boldsymbol{\omega}}_n^{(s)}\Vert^2.
\end{align}

Then, we accumulate the inequality (\ref{55}) over $s=0$ to $s=S$ as
\begin{align}
	&\frac{1}{S+1}\sum_{s=0}^S\mathbb{E}\left\{\Vert \nabla\tilde{\mathcal{L}}(\tilde{\boldsymbol{\omega}}^{(s)})\Vert^2\right\} \nonumber\\
	&\leq -\frac{2}{\eta(S+1)}\sum_{s=0}^S(\mathbb{E}\left\{\mathcal{L}(\boldsymbol{\omega}^{(s+1)})\right\}-\mathbb{E}\left\{\mathcal{L}(\boldsymbol{\omega}^{(s)})\right\}) \nonumber\\
	&+\frac{\eta A^2}{Z(S+1)}\sum_{s=0}^{S}\bigg(\eta^2L+\eta^2+\vert\sum_{n=1}^{N}a_n^{(s)}\phi_n\vert^2\bigg)\frac{1}{\sum_{n=1}^{N}a_n^{(s)}} \nonumber\\
	&+\frac{L^2}{(S+1)}\sum_{s=0}^{S}\frac{\sum_{n=1}^{N}a_n^{(s)}\mathbb{E}\left\{\Vert\boldsymbol{\omega}^{(s)}-\tilde{\boldsymbol{\omega}}_n^{(s)}\Vert^2\right\}}{\sum_{n=1}^{N}a_n^{(s)}},
\end{align}
where the third term captures the effect of model pruning and the second term shows the effect of generalization gap analysis. From Assumption \ref{Assump4}, the pruning effect term and the inequality is transferred as
\begin{align}
	&\frac{1}{S+1}\sum_{s=0}^S\mathbb{E}\left\{\Vert \nabla\tilde{\mathcal{L}}(\tilde{\boldsymbol{\omega}}^{(s)})\Vert^2\right\} \nonumber\\
	&\leq\frac{2(\mathcal{L}(\boldsymbol{\omega}^{(0)})-\mathcal{L}(\boldsymbol{\omega}^{*}))}{\eta(S+1)}+\frac{L^2B^2}{(S+1)}\sum_{s=0}^{S}\frac{\sum_{n=1}^{N}a_n^{(s)}\lambda_n^{(s)}}{\sum_{n=1}^{N}a_n^{(s)}} \nonumber\\
	&+\frac{\eta A^2}{Z(S+1)}\sum_{s=0}^{S}\bigg(\eta^2L+\eta^2+\vert\sum_{n=1}^{N}a_n^{(s)}\phi_n\vert^2\bigg)\frac{1}{\sum_{n=1}^{N}a_n^{(s)}} \nonumber\\
    &=\frac{2(\mathcal{L}(\boldsymbol{\omega}^{(0)})-\mathcal{L}(\boldsymbol{\omega}^{*}))}{\eta(S+1)}+\frac{\eta^3A^2(L+1)}{Z(S+1)}\sum_{s=0}^S\frac{1}{\sum_{n=1}^{N}a_n^{(s)}}\nonumber\\
    &+\sum_{s=0}^S\frac{\gamma_1\vert\sum_{n=1}^{N}a_n^{(s)}\phi_n\vert^2+\gamma_2\sum_{n=1}^{N}a_n^{(s)}\lambda_n^{(s)}}{\sum_{n=1}^{N}a_n^{(s)}},
\end{align}
where $\omega^{(*)}$ is the optimal model, $\gamma_1=\frac{\eta A^2}{Z(S+1)}$, and $\gamma_2=\frac{L^2B^2}{(S+1)}$. Finally, we obtain the average squared gradient norm, which is presented as the convergence rate of the FEEL model with model pruning and generalization analysis.

\bibliographystyle{IEEEtran}
\bibliography{reference.bib}
\end{document}